\documentclass[lettersize,journal]{IEEEtran}  

\IEEEoverridecommandlockouts                              

\usepackage{graphics} 
\usepackage{graphicx}
\usepackage{subfigure}
\usepackage{cool}
\usepackage{epsfig} 
\usepackage{amsmath} 
\usepackage{gensymb}
\title{\LARGE \bf
Design and Evaluation of a Compact 3D End-effector Assistive Robot for Adaptive Arm Support}

\author{Sibo Yang$^{1}$, Lincong Luo$^{2*}$, Wei Chuan Law$^{2}$, Youlong Wang$^{2}$,
Lei Li$^{2}$ and Wei Tech Ang$^{1,2}$
\thanks{This work was supported by the Rehabilitation Research Institute of Singapore (\textit{Corresponding author: Lincong Luo})}
\thanks{$^{1}$S. Yang and W.T. Ang are with the School of Mechanical and Aerospace Engineering, Nanyang Technological University, Singapore
        {\tt\small sibo001@e.ntu.edu.sg, wtang@ntu.edu.sg}} 
\thanks{$^{2}$ L. Luo, W.C. Law, Y. Wang, L. Li, and W.T. Ang are with Rehabilitation Research Institute of Singapore (RRIS), Nanyang Technological University, Singapore.
        {\tt\small albert.luo27@gmail.com}}%
}

\begin{document}

\maketitle
\thispagestyle{empty}
\pagestyle{empty}
\setlength{\intextsep}{3pt plus 1.0pt minus 1.0pt}

\begin{abstract}
We developed a 3D end-effector type of upper limb assistive robot, named as Assistive Robotic Arm Extender (ARAE), that provides transparency movement and adaptive arm support control to achieve home-based therapy and training in the real environment. The proposed system composes five degrees of freedom, including three active motors and two passive joints at the end-effector module. The core structure of the system is based on a parallel mechanism. The kinematic and dynamic modeling are illustrated in detail. The proposed adaptive arm support control framework calculates the compensated force based on the estimated human arm posture in 3D space. It firstly estimates human arm joint angles using two proposed methods: fixed torso and sagittal plane models without using external sensors such as IMUs, magnetic sensors, or depth cameras. The experiments were carried out to evaluate the performance of the two proposed angle estimation methods.
Then, the estimated human joint angles were input into the human upper limb dynamics model to derive the required support force generated by the robot. The muscular activities were measured to evaluate the effects of the proposed framework. The obvious reduction of muscular activities was exhibited when participants were tested with the ARAE under an adaptive arm gravity compensation control framework. The overall results suggest that the ARAE system, when combined with the proposed control framework, has the potential to offer adaptive arm support. This integration could enable effective training with Activities of Daily Living (ADLs) and interaction with real environments.
\end{abstract}

\section{INTRODUCTION}
Mobility impairment of the upper limbs, such as that caused by a stroke, can significantly impact activities of daily living (ADL) \cite{mayo2002activity, bos2016structured}. Rehabilitation or daily assistance plays an important role in improving the patient's quality of life. However, the conventional rehabilitation approach produces a high workload on the therapists and can not provide precise rehabilitation assessment. Therefore, the emergence of upper limb rehabilitation/assistive robotics enables task-oriented rehabilitation or ADL assistance, reducing time-consuming and improving the patient's quality of life \cite{maciejasz2014survey}. \par
\begin{figure}[thpb]
    \centering
    \subfigure[]{\label{exp_setup}
        \includegraphics[width=0.3\textwidth]{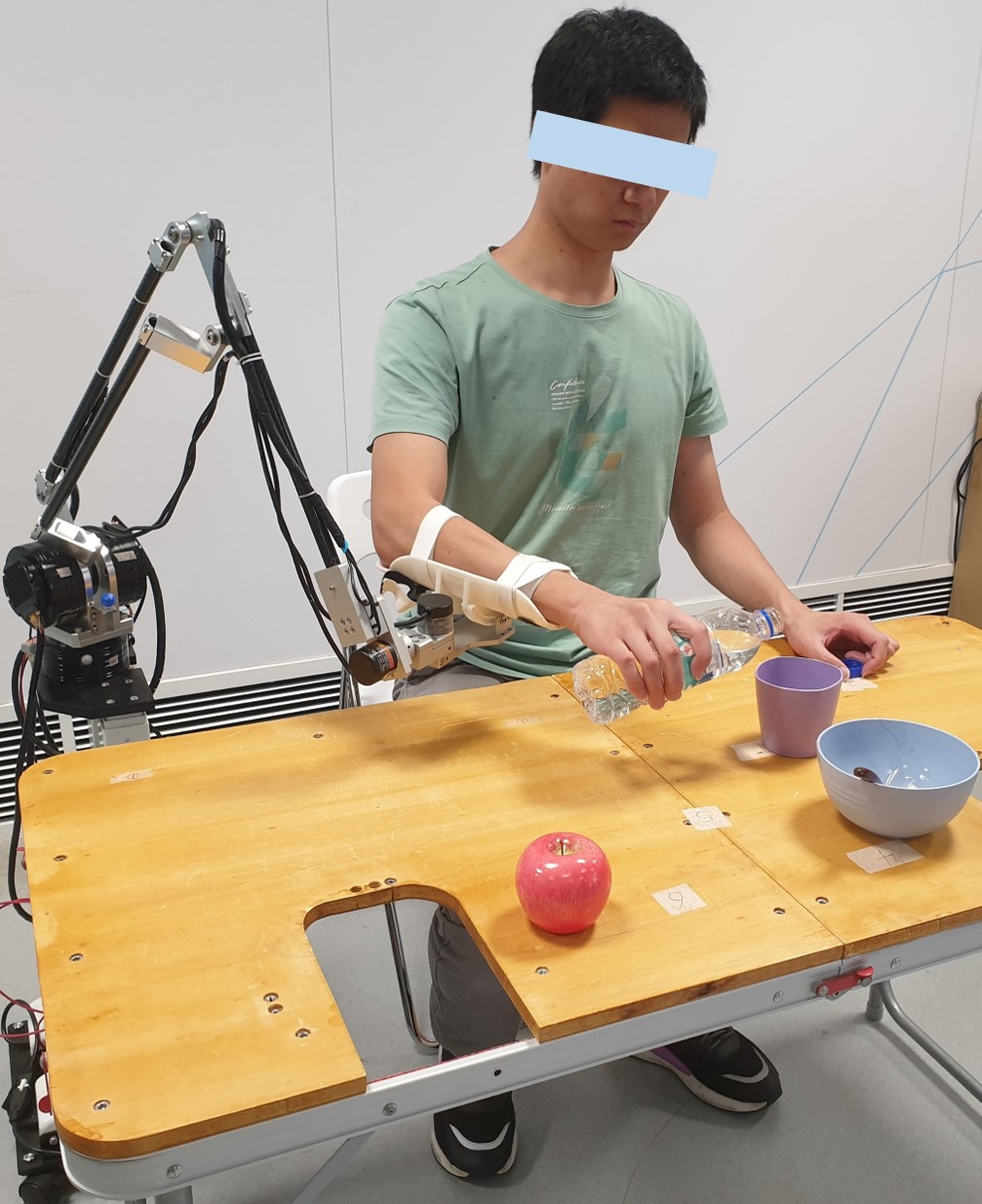}
        \label{fig:ADL}}
        
    \subfigure[]{\label{structure} 
        \includegraphics[width=0.45\textwidth]{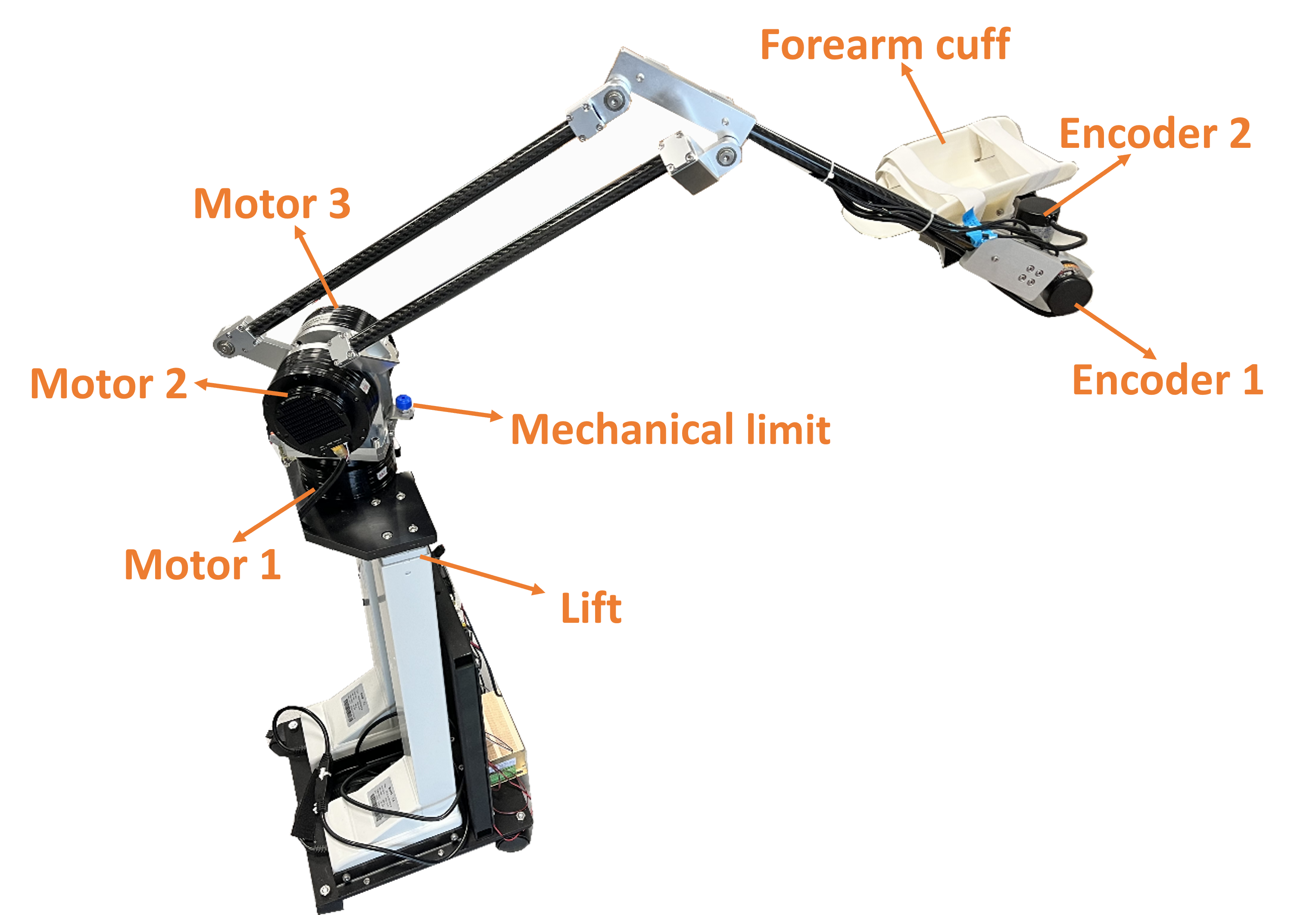}}%
    \caption{\textbf{Proposed system}. (a). A healthy subject is attached with ARAE prototype. (b). The mechanical structure of ARAE.}
    \label{fig:structure}
\end{figure}
Upper limb rehabilitation robots mainly include exoskeleton type and end-effector type \cite{qassim2020review}. Exoskeleton robots have been used for an extensive range of motion (ROM), including spatial 3D motions, which can fulfill most activities of daily life (ADL), for example, the ARMin \cite{nef2006armin} and Anyexo \cite{zimmermann2023anyexo}. However, robots of this kind are usually cumbersome and challenging to use at home. Most crucially, misalignment between the human and robotic joints is a common occurrence with this design. This misalignment can lead to unintended forces at the point of attachment, potentially resulting in additional harm to the patient's arm \cite{jarrasse2011connecting}. Modifying the lengths of the links to align with the joint kinematics of each patient also results in an increase in the time required for clinical setup. Furthermore, each of the joints is controlled by heavy motors, which add more mechanical inertia to the system. Hence, it can reduce the dynamic transparency of robots \cite{sun2021sensor}. In addition, the motor of exoskeleton robots needs to produce high torque to compensate for the human arm's load and mechanical components. Therefore, the motors with a high gear ratio are utilized in this type of robot, which diminishes the back-drivability of the robot \cite{matsuki2019bilateral}. In comparison, the end-effector robot did not suffer many of the mentioned drawbacks of exoskeletons. Such a robot refers typically to a two-DOF robot working in the planar plane \cite{luo2019greedy}, which interacts with the human forearm or wrist, such as MIT MANUS \cite{krebs2007robot}. However, this type of robot only covers fewer daily activities due to the limited working range in the 2D plane. Therefore, this approach is unsuitable for patients requiring training or assistance with complex Activities of Daily Living (ADL) tasks, particularly those involving three degrees of freedom (3D) movements. This limitation is addressed by 3D end-effector robots. \par
The 3D end-effector type of robot, such as EMU \cite{fong2017emu} and Burt \cite{duret2019robot}, is easy to set up and capable of operating within a large range of motion (ROM). However, one major issue with this type of robot is its bulky design, which limits its feasibility for home-based therapy. Additionally, to effectively support task-oriented rehabilitation and ADL assistance, the robot must provide satisfactory gravity compensation (GC) for the upper limb, as highlighted by \cite{kramer2007design}. Thus, the GC method can reduce the muscular effort of the user \cite{lobo2015adaptive} and enhance the transparency of movement during rehabilitation \cite{zimmermann2020towards}. Therefore, GC is crucial for patients with upper limb movement disorders \cite{luo2017design}. In addition, there is no correction for trajectory tracking, allowing the user to actively explore the range of motion during training of ADL \cite{dalla2021review}. A significant challenge with arm gravity compensation (GC) in 3D end-effector robots is that the required compensated force at the end-effector point changes depending on the arm's pose in three-dimensional space. In studies by \cite{fong2017emu} and \cite{crocher2018upper}, the researchers measured the angles of human joints during movements using magnetic sensors and then calculated the support force using the rigid link model and human dynamics model, respectively. However, the main limitation is that the joint angle estimation needs to be achieved by wearable magnetic sensors. This procedure can increase the complexity of use, and the magnetic sensor normally produces drifts or noises \cite{yun2007self}.\par
This paper introduces the Assistive Robotic Arm Extender (ARAE), designed to provide arm support in three-dimensional (3D) space for functional task training. The ARAE is capable of achieving high transparency in movement within 3D space and offers adaptive arm support based on estimated human postures. These features enable it to assist patients with Muscle Manual Testing (MMT) scores ranging from 1 to 4, as defined by \cite{cuthbert2007reliability}, in performing ADLs and interacting with real environments.  
The ARAE was designed using Quasi-direct drive motors, encoders, and a parallel mechanism, incorporating three active degrees of freedom (DOFs) and two passive DOFs. Utilizing the ARAE system, we also developed an adaptive control framework for compensating the gravity on the human arm, which calculates the compensatory force based on the estimated human postures. This framework comprises two main components. First, we introduce two modeling methods for estimating human joint angles: the fixed torso model (as proposed by our previous study \cite{yang2023adaptive}) and the sagittal plane models. These methods were evaluated through experiments involving reaching, placing, and drinking motions with four healthy subjects. To assess the accuracy of these angle estimation models, we conducted a comparative analysis between the joint angles projected by the models and the actual joint angles measured by a Motion Capture system (Mocap). Subsequently, the derived joint angles were used in the human dynamics model to calculate the arm support force. To validate the effectiveness of this system in reducing muscle energy in the assistive mode, electromyography (EMG) activities were measured on healthy subjects with and without the robot \cite{just2020human}. 
The remainder of the paper proceeds as follows: Section \ref{sec:design} describes the system of ARAE. Section \ref{sec:framework} illustrates the adaptive arm gravity compensation control framework. The experimental protocol and evaluation methods are presented in Section \ref{sec:exp}. In Section \ref{sec:results}, we demonstrate the validated results. Finally, Section \ref{sec:discussion} shows the discussion and \ref{sec:con} concludes the paper.\par
\section{Assistive Robotic Arm Extender System}
\label{sec:design}
\begin{figure}[!htbp]
  \centering
    \includegraphics[width=1\linewidth]{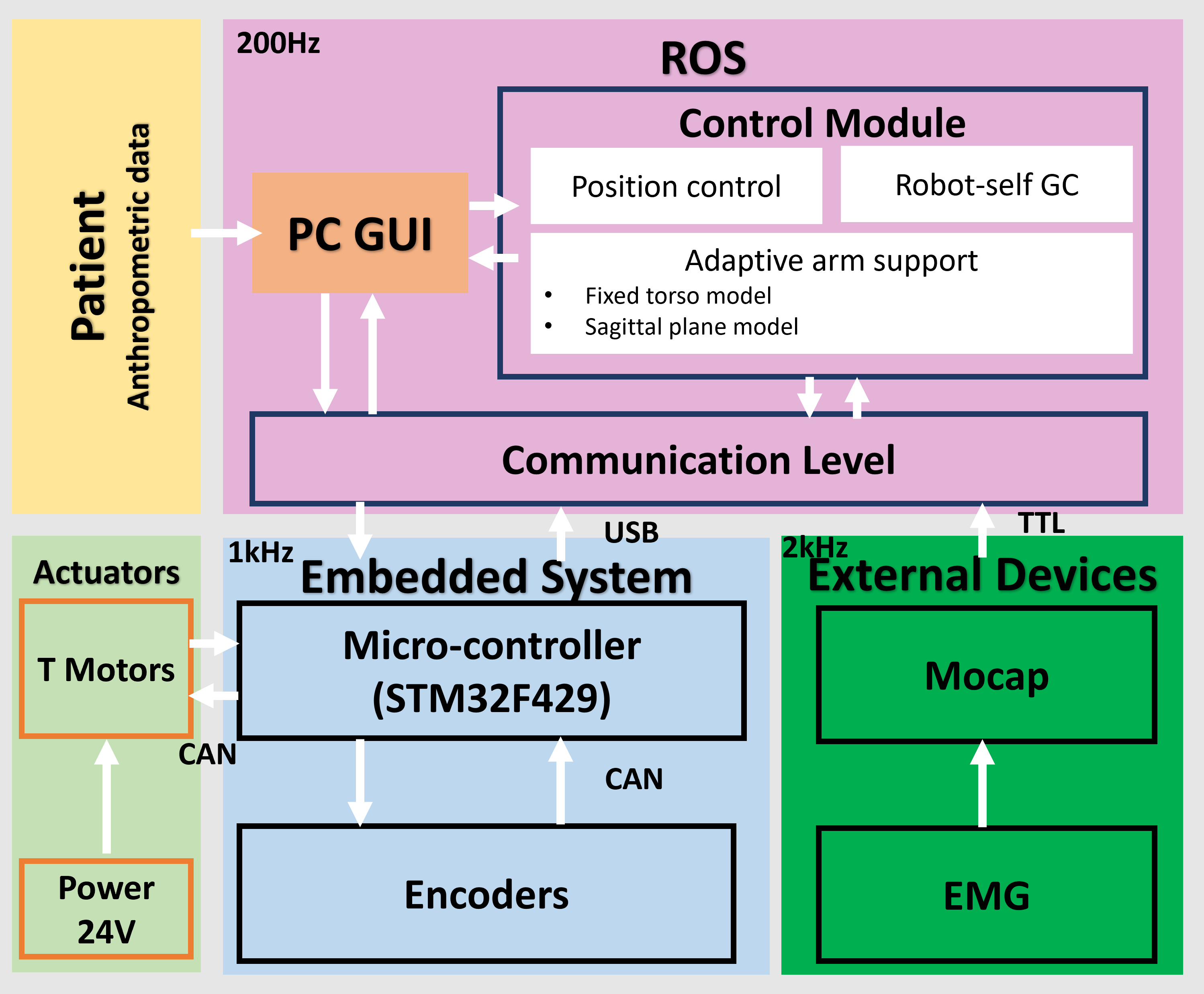}
  \caption{Flow chart of the robotic system}
  \label{fig:system overview}
\end{figure}

ARAE is an end-effector type of robot working in 3D space that can fulfill the working range of daily life activities. 
The comprehensive system encompasses the mechanical design of the robot, and places particular emphasis on the embedded and software system architecture, as shown in Fig. \ref{fig:system overview}. Following this, a dedicated subsection is devoted to elaborating on the robot's kinematic and dynamic models in detail. 
\subsection{Mechanical Design}
The invented robotic system consists of two sub-systems: a 3-DOF actuated robot arm and the end effector module with two passive joints, the overview of the system prototype is shown in Fig.\ref{fig:structure}.
\subsubsection{\textbf{Structure}} As depicted in Fig. \ref{fig:structure}, the base Motor $1$, mounted on the base plate, is engineered to enable rotary movement at the base. This is complemented by two additional motors, Motor $2$ and Motor $3$, positioned atop Motor $1$. Motor $2$ is responsible for driving Link 1, while Motor $3$ facilitates movement for Link 2. These components, along with Links 3 and 4, form a parallelogram mechanism, with Link 4 functioning as the output link that interfaces with the end-effector module. The blue mechanical limit serves as the protective limit for Motors $2$ and $3$.\par
The links, designed for high stiffness and minimal weight, are fabricated from carbon fiber tubes. This design strategy ensures a compact configuration, optimizing the transmission of actuator torque to the end user. The end-effector module attaches to Link 4, secured by clamping onto the link and an anti-rotation screw. 
For motion tracking and feedback, two encoders provide 2-DOF movement, with their output shafts serving as the rotation axes. The system's forearm cuff, designed to securely hold the user's forearm, also allows for slight rotational adjustment following the curve beneath the cuff. 
\subsubsection{\textbf{Actuation}}Achieving the mechanical adaptability of wearable robots, including assistive or rehabilitation robots, relies heavily on high-performance actuators. These actuators must meet specific criteria, including being lightweight, highly backdrivable, and possessing a high bandwidth. However, the state-of-the-art actuator like Series Elastic Actuator (SEA) needs to compromise the bandwidth to achieve high backdrivibility \cite{paine2013design}. Therefore, the mainstream use quasi-direct drive (QDD) motors to provide high torque and backdrivibility, which enables to promote the performance of physical human-robot interaction (pHRI) \cite{yu2020quasi}. Since ARAE's mechanical design is end-effector type, the main load of the motors is only the mechanical structure. We selected three QDD motors (T-Motor AK10-9 V2.0) used to provide 3-DOF movement. The selected motors enable to generate the peak torque of 48Nm. The maximum external load applicable at the end-effector is $12.43 kg$ when the parallel mechanism extends to its maximum working range. At this point, the end-effector load has the largest force arm relative to the motors. The peak loads at the end-effector can support the arm weight of $99\%$ of humans \cite{preedy2012handbook} and facilitate user interaction with real objects. This will allow for a variety range of human body and strength training. The maximum joint speed of 26 rad/s provided approximately 8 times higher speeds as required in ADL tasks. The back drive torque is 0.8Nm generating high performance of transparency during human-robot interaction (HRI). The safety limits are set for position, velocity, and torque control.\par

\subsection{Embedded and Software System Design}
An STM32F429 (STMicroelectronics) microcontroller (MCU) is utilized for the embedded system to communicate with the encoders and T-Motors through the CAN (Controller Area Network) bus while connecting with the Linux application through a serial port. 
The MCU communicates with the external ADC chip AD7606 through SPI (Serial Peripheral Interface). The sampling frequency of the encoder feedback and the low-level actuator control loop is fixed at 1kHz. A computer running Linux is used as a host PC for the data logging and user interface. The communication between MCU and the host PC is written in C++ as nodes for the robot operating system (ROS). The PC-based Graphical User Interface (GUI), developed in C++, facilitates real-time monitoring of the ARAE's operational status.

%
%

\subsection{Model Analysis}
\subsubsection{Forward kinematics model}
\label{sec:FK model}
The purpose of the forward kinematics model is to solve the end-effector position of ARAE with the five input joint angles $q_i$ ($i=1,2,3,4,5$). 
As shown in Fig.\ref{fig:ARAEwith human}, the \{$O_{R0}$\}:$O_{R0} - x_{0}y_{0}z_{0}$ is the robot base coordinate and the \{$O_{R5}$\}:$O_{R5} - x_{5}y_{5}z_{5}$ refers to the coordinate of end-effector point for robot. The Denavit-Hartenberg (D-H) algorithm is applied to derive the kinematics model.
The specific process of the forward kinematics model and the illustration of D-H parameters, including the length of the links, and the initial position of each joint are defined in the supplementary material S.I.\par
Since ARAE has only three active joints ($q_{1}, q_{2}, q_{3}$), the Jacobin matrix maps the first-order differential relationship between active joints and the position of joint 4 (modified end-effector position) ${ }^{R}\textbf{p}_{3}$ in Cartesian space. The robot jacobin matrix is $\textbf{J}_{R} \in \mathbb{R}^{3 \times 3}$, with detailed explanation provided in S.I.

\subsubsection{Inverse kinematics model}
The inverse kinematics (IK) model facilitates the achievement of a fully passive control mode for the ARAE. This allows the robot to maneuver the patient's arm to a predefined position without necessitating any muscular effort on the patient's part. \par
Given that the ARAE is equipped with two passive joints, denoted as \(q_{4}\) and \(q_{5}\), the general inverse kinematics model can yield an infinite set of solutions represented by \(q_{i}\). Therefore, the inverse kinematics (IK) model is constrained to only compute the active joint angles.
The specific expression of the inverse kinematics model is illustrated in the supplementary material S.II.
\subsubsection{Dynamic model of whole robotic structure}
The purpose of modeling robot dynamics is to achieve the gravity compensation of the mechanical structure. The gravity compensation feature of the ARAE ensures that the entire system operates in transparent mode, necessitating only minimal externally applied force. Furthermore, the robot is capable of maintaining a stable hover at the target position once the external force is withdrawn. The derived Euler–Lagrange dynamics in joint space are governed by the following Equation:
\begin{equation}
M_{R}(q_{i}) \ddot{q_{i}} + C_{R}(q_{i}, \dot{q_{i}}) \dot{q_{i}} + G_R(q_{i}) = \tau_{R}
\label{equ:D_model}
\end{equation}
where  $q_{i}, \dot{q_{i}}, \ddot{q_{i}} $ is limited to the three input joint angles ($i=1,2,3$) since there are only three active joint angles in ARAE. The $M_{R}(q_{i})$ represents the mass (inertia) matrix of robot, the $C_{R}(q_{i}, \dot{q_{i}})$ refers to the Coriolis and centripetal matrix, $G_R(q_{i})$ is the gravity vector and $\tau_{R}$ is the required joint vectors of three motors. Assuming the absence of inertia in the robot system, only the gravity term $G_R(q_{i})$ is considered for calculating the compensated joint torques.
\begin{figure}[thpb]
      \centering
      \includegraphics[width=0.48\textwidth]{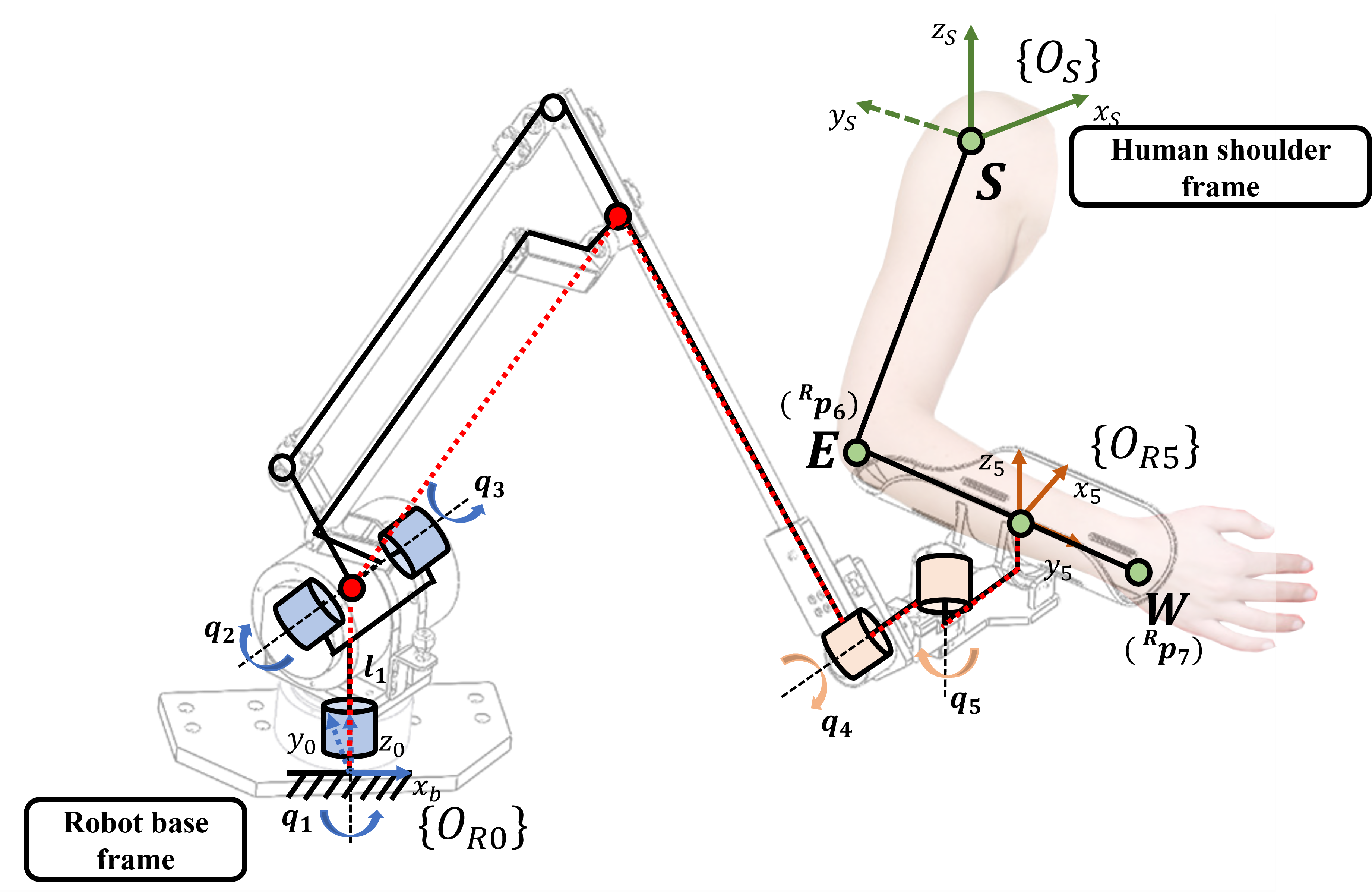}
      \caption{Schematic combined with human arm and ARAE robotic system. The human arm is modeled as a four degree of freedom link mechanism, including $3$ revolute joints at the shoulder joint and $1$ revolute joint at the elbow joint ($\textbf{E}$) under the human shoulder base coordinate {$O_{S}$}. The human shoulder base frame is denoted as \{$O_{S}$\}:$O_{S} - x_{S}y_{S}z_{S}$. $\boldsymbol{S}, \boldsymbol{E}$ and $\boldsymbol{W}$ represent the position vector of the shoulder joint, elbow joint, and wrist joint under the human shoulder base coordinate}
      \label{fig:ARAEwith human}
\end{figure}
\section{ADAPTIVE ARM GRAVITY COMPENSATION CONTROL FRAMEWORK}
\label{sec:framework}
\begin{figure*}[thpb]
\centering
    \includegraphics[width=0.95\textwidth]{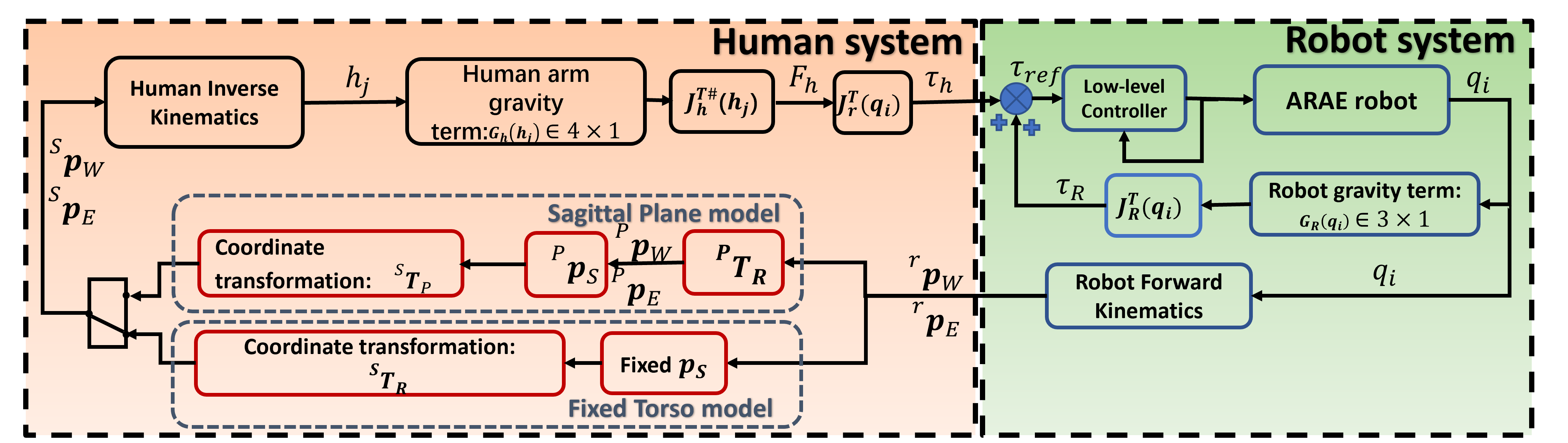}
    \caption{The system control framework demonstrates the interaction between the human and robot systems. The human system mainly refers to the proposed adaptive GC of the human arm control framework. Firstly, to estimate the human joint angles $h_j$ based on the $\textbf{P}_s, \textbf{P}_e$ and $\textbf{P}_w$ obtaining from the fixed torso model or sagittal plane model. Secondly, calculate the human arm needed support force $F_h$, then transmit to the torque provided by the robot $\tau_{h}$. The reference torque $\tau_{ref}$ is fed into the motor controller of robot system $\tau_{ref} = \tau_c + \tau_R$. $\tau_R$ is the gravity torque of the robot structure. }  
\label{fig:control diagram}
\end{figure*}

Adaptive arm gravity compensation means that the support force provided by the robot at the end-effector can change with the arm posture. As shown in Fig.\ref{fig:control diagram}, the entire control framework reflects the interaction between the human-robot system. The proposed adaptive gravity compensation of the human arm is represented by the joint angle estimation method and calculation of the human required support force.
\subsection{Human joint angle estimation method}
\label{sec:angle estimation}
Since the human forearm is securely fastened to the cuff using Velcro, it is feasible to align the positions of the elbow and wrist joints with the two endpoints of the cuff. As depicted in Fig. \ref{fig:ARAEwith human}, the elbow and wrist points of the subject are estimated to coincide with ${ }^{R}\textbf{p}_{6}$ and ${ }^{R}\textbf{p}_{7}$ (mentioned in the supplementary material S.I), respectively, as expressed in the following manner: 
\begin{equation}
{ }^{R}\textbf{p}_{E} = { }^{R}\textbf{p}_{6}
\label{equ:elbow}
\end{equation}
\begin{equation}
{ }^{R}\textbf{p}_{W} = { }^{R}\textbf{p}_{7}
\label{equ:wrist}
\end{equation}
where ${ }^{R}\textbf{p}_{E}$ and ${ }^{R}\textbf{p}_{W}$ are the human elbow and wrist positions in robot base coordinate.


\subsubsection{\textbf{Fixed torso model}}
\label{subsec:fixed}
In this model, human torso movement is constrained to a fixed position, allowing the shoulder point $S$ to be assumed as stationary. After requiring the elbow and wrist position under robot based frame, the ${ }^{R}\textbf{p}_{E}$ and ${ }^{R}\textbf{p}_{W}$ are transferred under the human shoulder base frame \{$O_{S}$\}:$O_{S} - x_{S}y_{S}z_{S}$, as shown in the following equation. 
\begin{equation}
\label{eq:phe}
\left[{ }^{S}\textbf{p}_{E}, { }^{S}\textbf{p}_{W}\right] = { }^{S}\textbf{T}_{R} \cdot[{ }^{R}\textbf{p}_{E}, { }^{R}\textbf{p}_{W}]
\end{equation}
\begin{equation}
{ }^{S}\textbf{T}_{R}=\left[\begin{array}{cccc}
\cos (\psi) & -\sin (\psi) & 0 & x_{SR} \\
\sin (\psi) & \cos (\psi) & 0 & y_{SR} \\
0 & 0 & 1 & z_{SR} \\
0 & 0 & 0 & 1
\end{array}\right]
\end{equation}
where ${ }^{S}\textbf{T}_{R}$ is the transformation matrix from the robot base frame to the human shoulder base frame. 
The rotational angle $\psi$ equal to $-\frac{\pi}{2}$. The $x_{SR},y_{SR},z_{SR}$ represent the position of the original point of the robot base coordinate in the fixed human shoulder frame. These values must be entered into the PC GUI as initial parameters. The ${ }^{S}\textbf{p}_{E}$ and ${ }^{S}\textbf{p}_{W}$ are referred to as elbow position and wrist position under the fixed human shoulder frame. \par 
Then, we use the human inverse kinematics model to derive the human joint angles $h_j$, including shoulder abduction/adduction ($h_1$), shoulder flexion/extension ($h_2$), shoulder internal/external rotation ($h_3$), and elbow flexion/extension ($h_4$). The established human inverse kinematics model is given by:
\begin{equation}
\begin{aligned}
\centering
\label{equ:HIK}
& h_j = \textbf{IK} ({ }^{S}\textbf{p}_{E}, { }^{S}\textbf{p}_{W}, l_{U_{cal}}, l_F)\ \quad j \in\{1,2,3,4\} \\
& l_{U_{cal}} = \| { }^{S}\textbf{p}_{E} \| \\
\end{aligned}
\end{equation}
where the ${ }^{S}\textbf{p}_{E}$ and ${ }^{S}\textbf{p}_{W}$ are the position vectors input to the human IK model. The $l_F$ represents the length of the forearm, defined as an initial parameter subject to anthropometric data. The detailed expression of the human arm inverse kinematics model is demonstrated in our previous study \cite{yang2023adaptive}. We assume that the original point of the shoulder frame under the human shoulder frame is denoted as $\textbf{p}_{S}$, and it is the fixed position. Therefore, the length from ${ }^{S}\textbf{p}_{E}$ to the fixed shoulder point is calculated as $l_{U_{cal}}$ rather than directly using the actual human upper arm length ($l_U$) due to the change of derived ${ }^{S}\textbf{p}_{E}$.

However, a primary limitation of this model is its assumption that the shoulder position remains fixed. In practical scenarios, the shoulder position typically shifts in conjunction with torso movements, particularly during actions like reaching for distant positions. To address this, we introduce an enhanced model, which we refer to as the 'Sagittal plane model'. 
\subsubsection{\textbf{Sagittal plane model}}
\begin{figure}[thpb]
      \centering
      \includegraphics[scale=0.38]{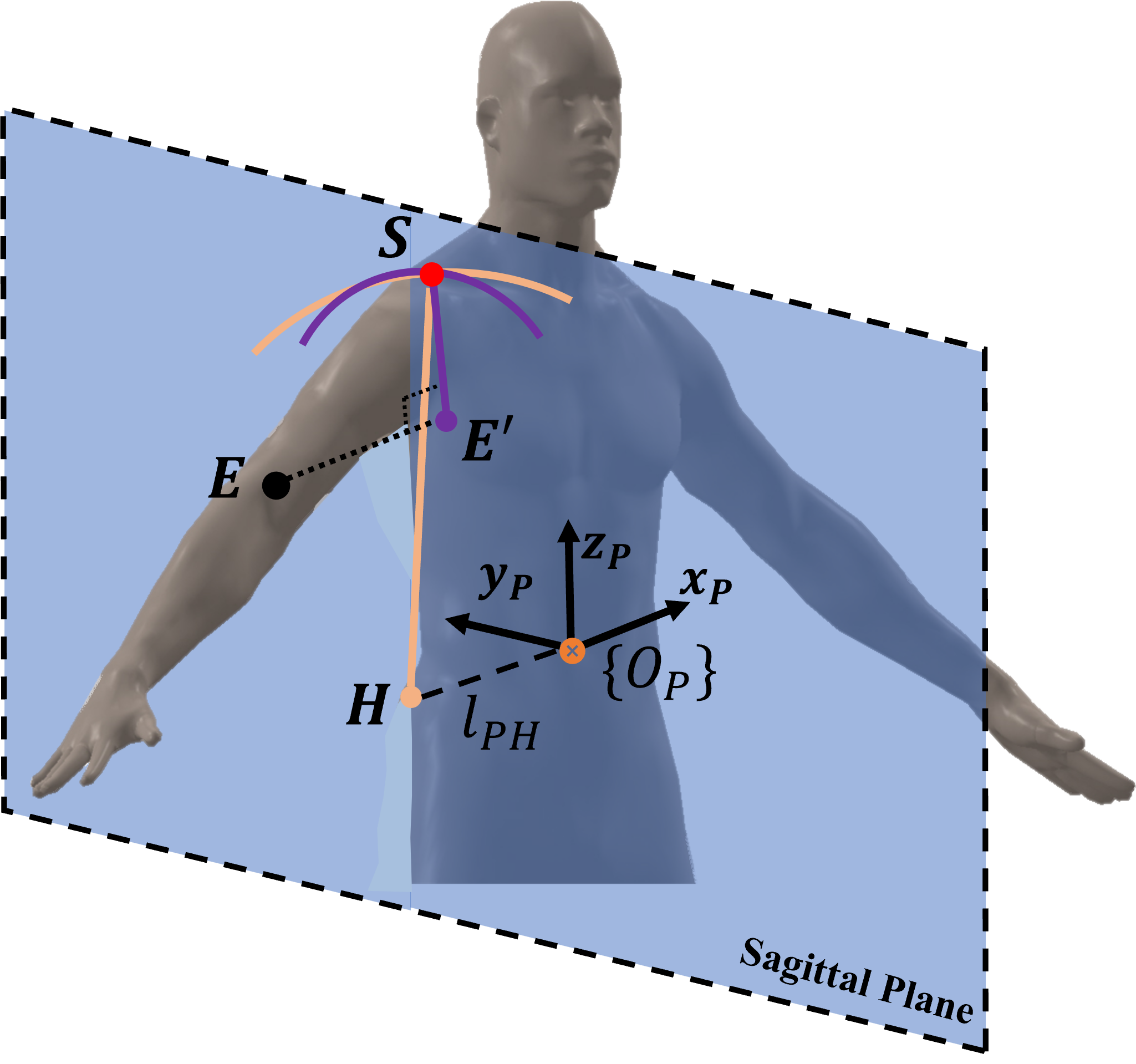}
      \caption{Schematic diagram for obtaining the shoulder position. The COM of the pelvis is the original point of the pelvis base coordinate \{$O_{hp}$\}:$O_{hp} - x_{hp}y_{hp}z_{hp}$. $\textbf{E'}$ is the projection of the elbow joint in the sagittal plane. $H$ represents the hip joint which is located at the sagittal plane and $x_P$ axis}
      \label{fig:sagittal}
\end{figure}

In this model, the shoulder position $\textbf{p}_S$  is assumed to move within the sagittal plane of the human torso. Due to the movement of the shoulder position, establishing the human base coordinate at the shoulder as a fixed reference frame becomes impractical. Therefore, it is advisable to relocate the human base frame to the center of the pelvis. This approach allows the assumption that the original point of the human pelvis base coordinate, denoted as \{$O_{P}$\}:$O_{P} - x_{P}y_{P}z_{P}$. This position remains relatively fixed, especially when the user is seated, as illustrated in Figure \ref{fig:sagittal}. Despite potential torso movements in various directions, this assumption provides a stable reference point. \par
As mentioned above, the elbow position and wrist position (${ }^{R}\textbf{p}_{E}$ and ${ }^{R}\textbf{p}_{W}$) can be derived from the robot forward kinematics model through Equation \ref{equ:elbow} and \ref{equ:wrist}. Unlike the fixed torso model, the ${ }^{R}\textbf{p}_{E}$ and ${ }^{R}\textbf{p}_{W}$ are transferred to the human-based pelvis frame, as shown in the following equation.
\begin{equation}
\label{eq:phe}
\left[{ }^{P}\textbf{p}_{E}, { }^{P}\textbf{p}_{W}\right] = { }^{P}\textbf{T}_{R} \cdot [{ }^{R}\textbf{p}_{E}, { }^{R}\textbf{p}_{W}]
\end{equation}
\begin{equation}
{ }^{P}\textbf{T}_{R}=\left[\begin{array}{cccc}
\cos (\psi) & -\sin (\psi) & 0 & x_{PR} \\
\sin (\psi) & \cos (\psi) & 0 & y_{PR} \\
0 & 0 & 1 & z_{PR} \\
0 & 0 & 0 & 1
\end{array}\right]
\end{equation}
Where ${ }^{P}\textbf{T}_{R}$ refers to the matrix from the robot frame to the human pelvis frame. $\psi$ equal to $-\frac{\pi}{2}$. The $x_{PR}, y_{PR}, z_{PR}$ represent the position of the original point of the robot base coordinate in the fixed human pelvis frame. These values also must be entered into the PC GUI as initial parameters. The ${ }^{P}\textbf{p}_{E}$ and ${ }^{P}\textbf{p}_{W}$ are the elbow and wrist positions under the pelvis coordinate.\par
We established the geometric model in the sagittal plane. As shown in Figure. \ref{fig:sagittal}, the hip joint position $H$ is assumed to be located on the sagittal plane and $x_P$ axis. $l_{SH}$ refers to the initial parameter from the hip to the shoulder. This parameter is an anthropometric value, unique to each user's torso length. $l_{PH}$ is another parameter referring to half of the torso width specific to different subjects. Thus, there are two arcs formed in the sagittal plane. One is an arc with the hip as the center of the circle and $l_{SH}$ as the radius. The other is the arc with the projected elbow joint $\textit{E'}$ as the center of the circle and $l_{SE'}$ as the radius. Therefore, the intersection point of two arcs is the derived shoulder position under pelvis coordinate, denoted by ${ }^{P}\textbf{p}_{S} = \left[{ }^{P}\textbf{p}_{S}(x), { }^{P}\textbf{p}_{S}(y), { }^{P}\textbf{p}_{S}(z)\right]$.
The two constrained problems are written as follows:
\begin{equation}
\begin{array}{ll}
& \left\|{ }^{P}\textbf{p}_{S}-{ }^{P}\textbf{p}_{H}\right\|^2=l_{S H}^2 \\
& \left\|{ }^{P}\textbf{p}_{S}-{ }^{P}\textbf{p}_{E'}\right\|^2=l_{SE'}^2 \\
\text{where} & l_{SE'} = \sqrt{l_{U}^2 - \left({ }^{P}\textbf{p}_{S}(x) - l_{PH}\right)^2}\\
\end{array}
\label{eq:arcs}
\end{equation}

Where the hip joint position is denoted as ${ }^{P}\textbf{p}_{H} = \left[-l_{PH}, 0, 0\right]$ and $l_U$ represents the length of the upper limb. Subsequently, the elbow and wrist positions under the pelvis frame must be transferred to the derived shoulder frame, which can be denoted as:
\begin{equation}
\label{eq:p2s}
\left[{ }^{S}\textbf{p}_{E}, { }^{S}\textbf{p}_{W}\right] = { }^{S}\textbf{T}_{P}\cdot[{ }^{P}\textbf{p}_{E}, { }^{P}\textbf{p}_{W}]
\end{equation}
\begin{equation}
{ }^{S}\textbf{T}_{P}=\left[\begin{array}{cccc}
1 & 0 & 0 & { }^{P}\textbf{p}_{S}(x) \\
0 & 1 & 0 & { }^{P}\textbf{p}_{S}(y) \\
0 & 0 & 1 & { }^{P}\textbf{p}_{S}(z) \\
0 & 0 & 0 & 1
\end{array}\right]
\end{equation}
where ${ }^{S}\textbf{T}_{P}$ refers to the homogeneous transformation matrix from fixed pelvis frame to derived shoulder frame. \par
Finally, the ${ }^{S}\textbf{p}_{E}$ and ${ }^{S}\textbf{p}_{W}$ are input to the human arm inverse kinematics model, as shown by Equation. \ref{equ:HIK}. 
\subsection{Arm gravity compensation strategy}
After estimating the joint angles of the human arm, the required support force can be computed using the human arm dynamics model. The detailed explanation is illustrated in the supplementary material S.III. As shown in Figure. \ref{fig:ARAEwith human}, the human arm model is modeled as a link mechanism with four degrees of freedom. The center of mass of these two links is shown as $m_U$ and $m_L$. \par
The required force of the end-effector to support the human arm's weight can then be calculated using the human arm model as follows:
\begin{equation}
    \mathbf{F}_h = \mathbf{J}_h^{T \#}\left(\mathbf{h}_j\right) \mathbf{G}_h\left(\mathbf{h}_j\right)
\label{equ:hd2}
\end{equation}
where $\mathbf{h}_j$ is the estimated human joint angles. The $\mathbf{J}_h^{T \#}\left(\mathbf{h}_j\right)$ is the pseudo-inverse of $\mathbf{J}_h^T(\mathbf{h}_j)\in \mathbb{R}^{3 \times 4}$ and $\mathbf{G}_h$ is the gravity term of human arm. The detailed expressions are illustrated in the supplementary material S. III.  The calculated force $\mathbf{F}_h$ varies in both magnitude and direction across the workspace, depending on the human arm joint angles $\mathbf{h}_j$.\par
Subsequently, the calculated force $\mathbf{F}_h$ is mapped to the compensated torque for human arm $\boldsymbol{\tau}_{h}$ in robot joint space as:
\begin{equation}
\boldsymbol{\tau}_{h}=J_R^{\mathrm{T}}\left(\mathbf{q}_i\right) \mathbf{F}_h
\end{equation}
The resultant reference torque $\boldsymbol{\tau}_{ref}$ to be provided by the active motors is as follows:
\begin{equation}
\boldsymbol{\tau}_{ref}=\boldsymbol{\tau}_{h} + \boldsymbol{\tau}_{R}
\end{equation}
where $\boldsymbol{\tau}_{R}$ was derived from the robot dynamics Equation. \ref{equ:D_model}.

\section{EXPERIMENTAL EVALUATIONS}
\label{sec:exp}
To evaluate the proposed adaptive gravity compensation framework on the ARAE robot, we first verified the proposed human joint angle estimation methods. Then, the effects of adaptive arm support force were evaluated by surface Electromyography (sEMG). 
\subsection{Subject Information and Initial Calibration}
\label{subsec:cal}
The Institutional Review Board of Nanyang Technological University (IRB-2022-821) approves the experimental protocol. After reviewing the informed consent form, four right-handed healthy subjects ($4$ males, $29\pm 2$ years old) were involved in the experiments. The mean mass of the participants is $75.05\pm2.5 kg$ and the mean height is $178\pm 4.23cm$. The mean upper limb length ($l_U$) is $29.91\pm0.25 cm$ and the forearm length ($l_F$) is $26.43\pm0.66 cm$. Moreover, the mean of trunk length ($l_{SH}$) is $38.50\pm1.04 cm$ and the mean of trunk width ($l_{PH}$) is $17.93\pm0.64 cm$. \par 
Before the official experiments began, a calibration trial was performed to measure the initial parameters by the Mocap system, including the translation distance from the shoulder joint to the robot base ($x_{SR}, y_{SR}, z_{SR}$) and from the COM of the torso to robot base ($x_{PR}, y_{PR}, z_{PR}$), respectively. Additionally, kinematic parameters of the human body, such as arm length, trunk length, and trunk width in the upright position, were measured. The subjects were instructed to wear the robot and then sit at the table, maintaining an upright and stable torso posture, as depicted in Figure \ref{exp_setup}. During this setup, both the ARAE and the Mocap system recorded data for 10 seconds. The initial shoulder joint position was recorded under the mocap-based coordinate.\par
\subsection{Experiment Protocol for Evaluating Angle Estimation Method}
\begin{figure}[thpb]
      \centering
    \includegraphics[width=0.4\textwidth]{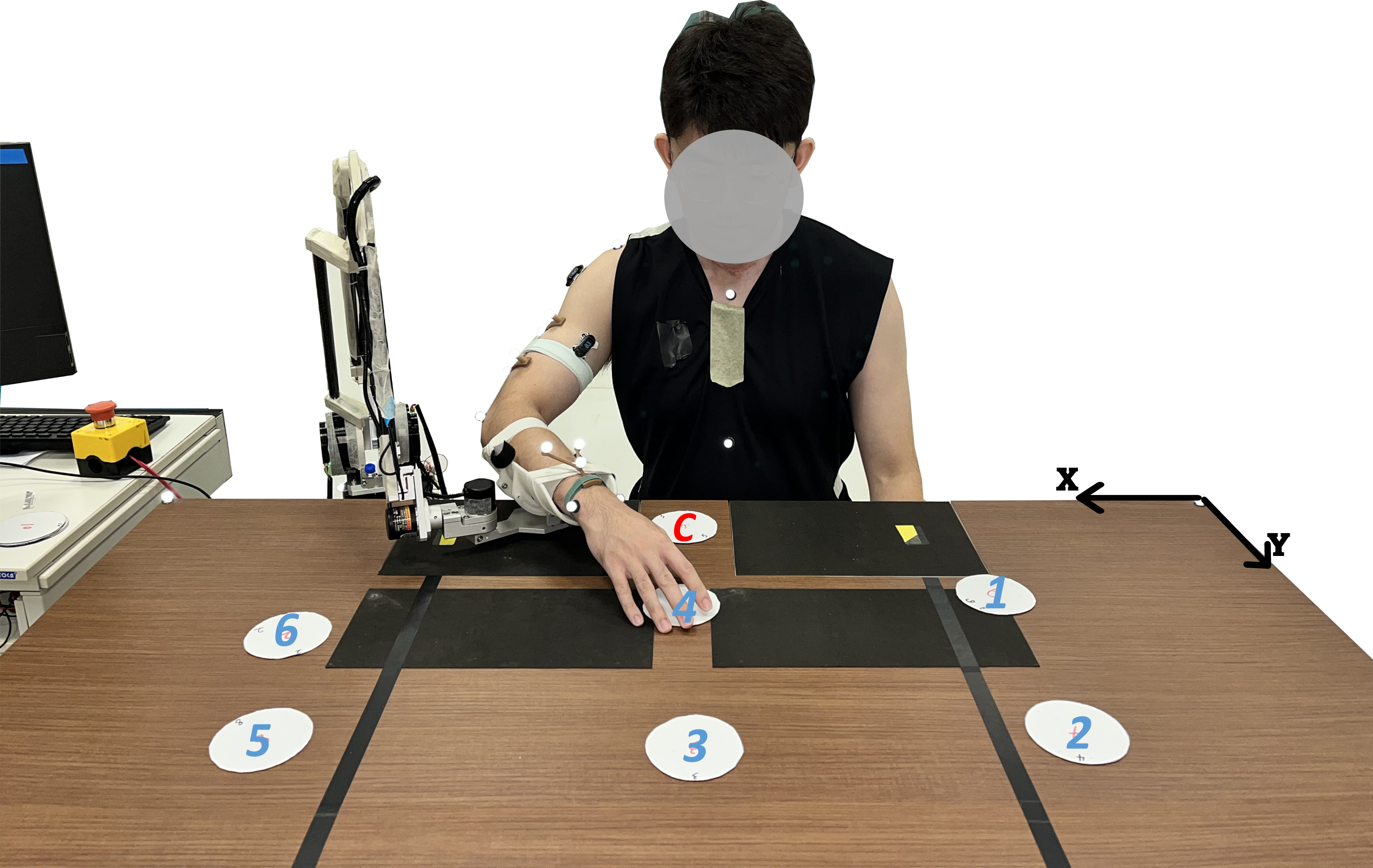}
    \caption{Proposed system for evaluation; (a). The subject wears the ARAE robot and sits in the capture volume of the Mocap system. The subject sits in front of a table and attaches the forearm to the ARAE robot. There are six labeled positions (Blue) and one original point (Red) located at the experimental table}
\label{fig:exp_table}
\end{figure}

The first experiment (Exp1) was conducted to evaluate the human joint angle estimation methods by the Mocap system. Six pre-defined positions are labeled on the table, shown in Fig.\ref{fig:exp_table}. The starting position is marked by the circular label that is closest to the human body. After each set of motions, the original position is set for resting the arm. The label $3$ is the farthest position from the body, making the subject do trunk compensation movements in the sagittal plane. The rest of the labels are located in a square. All subjects performed five trials for six labeled positions. Each trial involved continuous movements, specifically reaching and drinking activities performed with a real cylinder, effectively simulating ADL (Activities of Daily Living) task training. The detailed procedures are illustrated as follows:
\begin{enumerate}
    \item Firstly, the subject's arm starts by moving from the original position to the instructed label.
    \item After touching the labeled position, the subject lifts the cylinder and places it on their mouth.
    \item Finally, the subject brings their hand back to the labeled position and then returns to the original position.
\end{enumerate}
When the subjects move towards distal positions (Label 3), the subjects need to move the torso in the sagittal plane, which can mimic the trunk compensation movement. \par
\subsection{Experimental Protocol for Evaluating Effects on Human Arm}
The second experiment (Exp2) was to assess the impact of the adaptive arm support control framework on the human arm by measuring muscle activity using sEMG. To be able to independently verify the effect of the proposed adaptive control method on the human arm, the subjects did not take a physical object in this experiment. Exp2 was divided into three experimental sessions based on the different auxiliary control modes (No Robot and With Robot). In Exp2-1 (No Robot mode), the subject did the tasks without wearing the robot. Meanwhile, Exp2-2 and Exp2-3 employed the adaptive arm gravity compensation framework as the control mode. Specifically, the angle estimation method for Exp2-2 is based on the fixed torso model, whereas Exp2-3 utilizes the sagittal plane model. Each of the experimental sessions included three tasks, illustrated as follows: 
\begin{enumerate}
    \item Forward Reaching (FR): move the hand from the original position to label 3 position
    \item Lateral Reaching (LR): move the hand from the original position to label 2, then label 5 position
    \item Hand to Mouth (H2M): move the hand from the original position to the mouth
\end{enumerate}

\subsection{Evaluation Methods}
\subsubsection{Kinematic Data Preprocessing}
\begin{figure}[thpb]
\centering
\subfigure[]{\label{side_view} \includegraphics[width=0.18\textwidth]{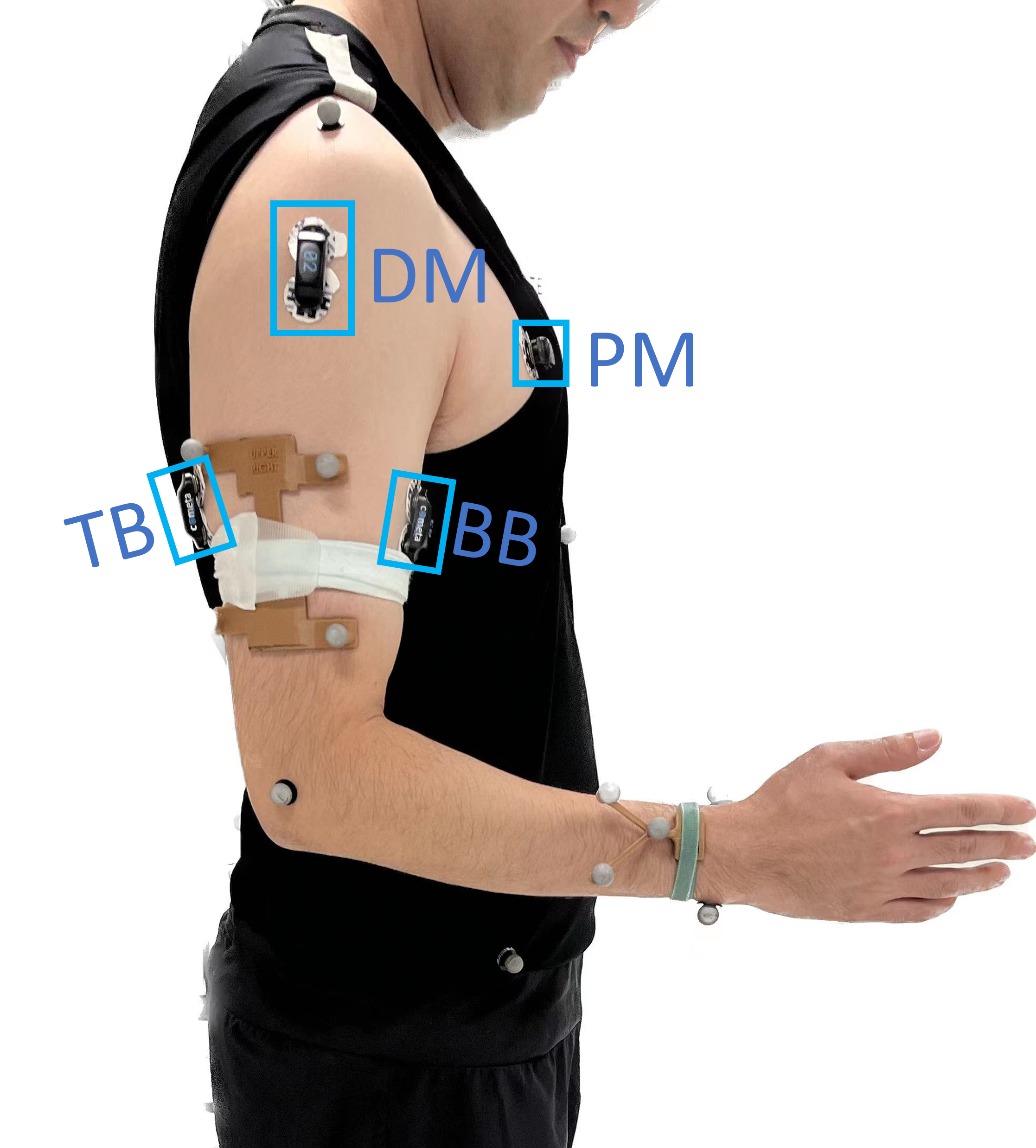}}
\hfill
\subfigure[]{\label{back_view} \includegraphics[width=0.15\textwidth]{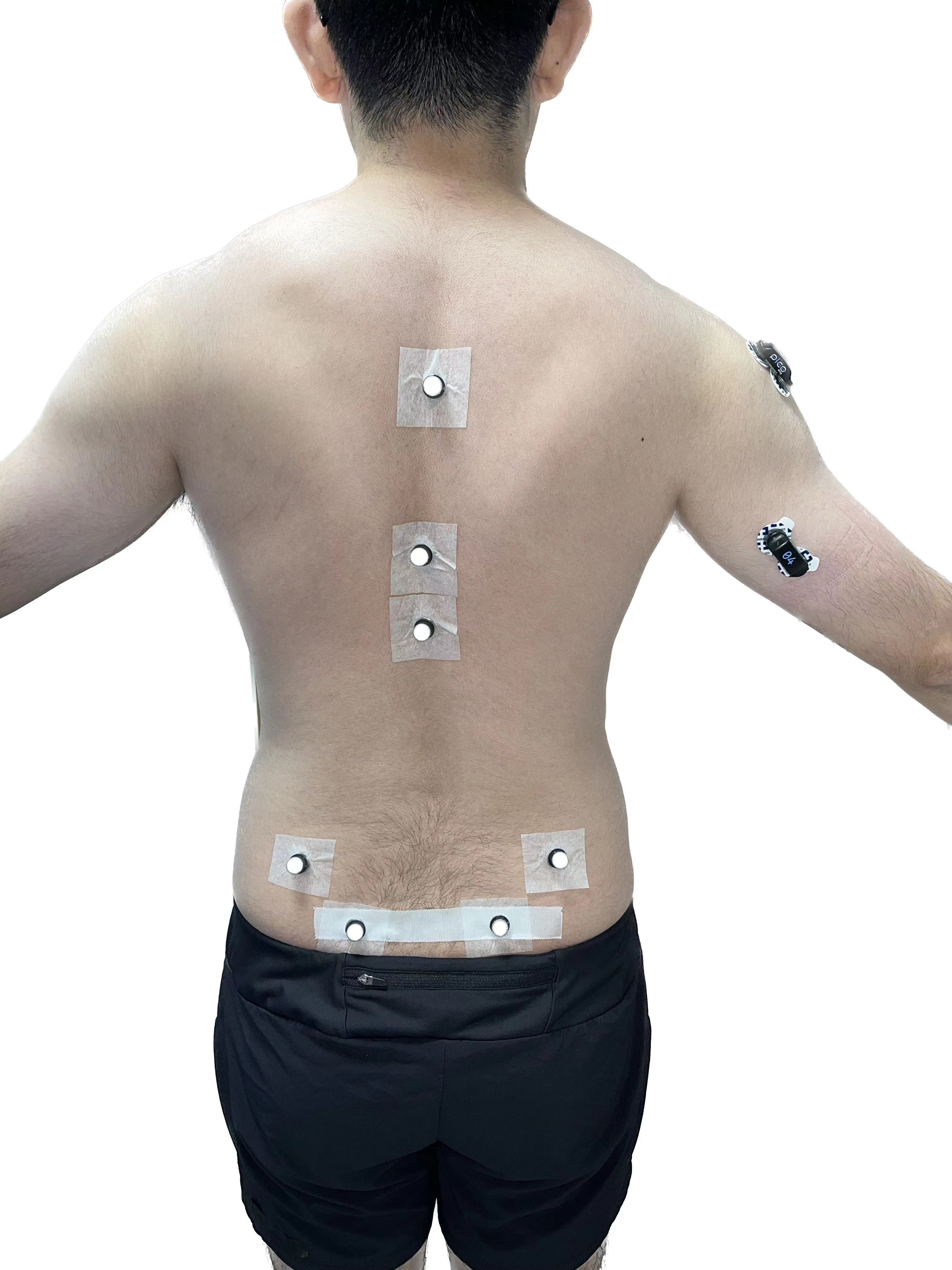}}%
\caption{The placement of sEMG and makers on the participant. (a) is the side view and (b) is the back view.}
\label{fig:sensors}
\end{figure}

To verify the proposed human joint angle estimation methods, the kinematic joint angles of the human arm were collected using the Qualisys Mique M3 motion capture (Mocap) system as the ground truth. The Mocap system includes 18 Qualisys A12MP cameras and a Qualisys task manager, an interface for managing the capture sessions and exporting the data at 200 Hz. The retro-reflective markers were placed on each healthy subject's body, including the thoracic spinous, right and left anterior superior iliac spines, sternum, upper arm, and forearm clusters, as shown in Fig. \ref{fig:sensors}. Since each subject wore a sleeveless T-shirt, the circular magnets were used to ensure that the position of the markers would not change for those markers that were covered by clothing at the torso. The system synchronization between the Mocap and ARAE robot is done by a DAQ board.\par
The markers' locations recorded by the Mocap system were sampled at 200 Hz. Then, Visual3D - a professional software- was used to transfer the marker location into human joint angles and each joint position under the Mocap world frame. Furthermore, the PC logged the corresponding data of the motors and encoders of the ARAE robot at 100 Hz. After the experiments, all the collected data were off-lined and analyzed by MATLAB R2022a. The kinematic data from Visual3D were downsampled to 100 Hz, which can synchronize with the measured data from the robot.\par
\subsubsection{sEMG Data Preprocessing}
To evaluate the adaptive arm gravity compensation framework of ARAE on the human body, muscle activities were recorded by the wireless sEMG system (Cometa Picolite, ITALY) at 2000 Hz. As shown in Fig. \ref{fig:sensors}, the EMG electrodes were placed on four upper limb muscles, including Pectoralis Major (PM), Deltoid Medial (DM), Bicep Brachii (BB), and Triceps Brachii (TB) following SENIAM guidelines \cite{hermens2000development}. At the beginning of each session, each subject needs to do Maximum Voluntary Contraction (MVC) which was later used to normalize the EMG signals. The recorded data were offline processing which involved two stages of notch filtering (using an IIR notch filter with a cutoff frequency of 50Hz to eliminate powerline interference and another at 1.67Hz to remove heartbeat noise); it was then subjected to high-pass filtering (via a 10th order Butterworth filter with a 20Hz cutoff frequency); the data was rectified by computing the absolute value; and subsequently smoothed with a low-pass filter (a 10th order Butterworth filter at a 4Hz cutoff frequency), as followed by \cite{clancy2016single}. All data from the sEMG channels were synchronized with the Mocap and ARAE systems through the DAQ board. 
\subsubsection{Metrics for Angle Estimation Methods}
The Mean absolute error (MAE) was used to evaluate the performance of the proposed angle estimation methods, which was denoted as 
\begin{equation}
MAE = \left|\frac{\sum_{i=1}^n \theta_j-h_j}{n}\right|
\label{equ:MAE}
\end{equation}
where $\theta_j$ is the ground truth human joint angles captured by the Mocap system and $h_j$ is the derived human joint angles using proposed models. $n$ represents the total number of data frames in each trial. 
\subsubsection{Metric for Adaptive Support Force}
The EMG data was filtered, and then the mean averaged value (MAV) was calculated for each EMG channel in reference to each task of Experiment 2. The percent change of the MAV ($\Delta$MAV\%) was conducted to measure the decrease in muscle activities from the control mode of Exp2-1 to Exp2-2 and Exp2-3, respectively. The $\Delta$MAV\% can be expressed as:
\begin{equation}
\Delta MAV\% = \frac{MAV_{Exp2-i} - MAV_{Exp2-1}}{MAV_{Exp2-1}}\cdot 100\%  \quad i \in\{2,3\}
\end{equation}

\section{Results}
\label{sec:results}
\subsection{Verification of Joint Angle Estimation}
To verify the two proposed human joint angle estimation methods, we input the robot joint angles into the derived model and output the estimated human joint angles, including shoulder abduction/adduction (SA), shoulder flexion/extension (SF), shoulder internal/external rotation (SR), and elbow flexion/extension (EF).
\subsubsection{Overall analysis}
Regarding the overall analysis, the types of joint angles were not classified during the analysis. As shown in Tab.\ref{tab:overall}, the mean absolute error (MAE) was calculated for all trials, and this value averages the MAE of the four types of joint angles. The performance of the sagittal plane model is slightly better than the fixed torso model, which obtains 5.37\degree. However, there was no significant difference in the accuracy of the mean joint angles MAE ($p=0.54$ Wilcoxon sign rank test). \par
\begin{table}[h]
\centering
\caption{Mean Absolute Error for subjects}
\label{tab:overall}
\begin{tabular}{ccc}
\hline
\multicolumn{3}{c}{MAE for different subjects (unit: degree)} \\ \hline
Subject & Fixed torso model & Sagittal plane model \\ \hline
1       & 5.70              & 5.03                 \\
2       & 5.59              & 5.92                 \\
3       & 5.20              & 4.52                 \\
4       & 5.99              & 6.01                 \\ \hline
Mean    & 5.62              & \textbf{5.37}        \\ \hline 
\end{tabular}
\end{table}
\subsubsection{Local analysis}
\begin{figure}[thpb]
      \centering
      \includegraphics[scale=0.42]{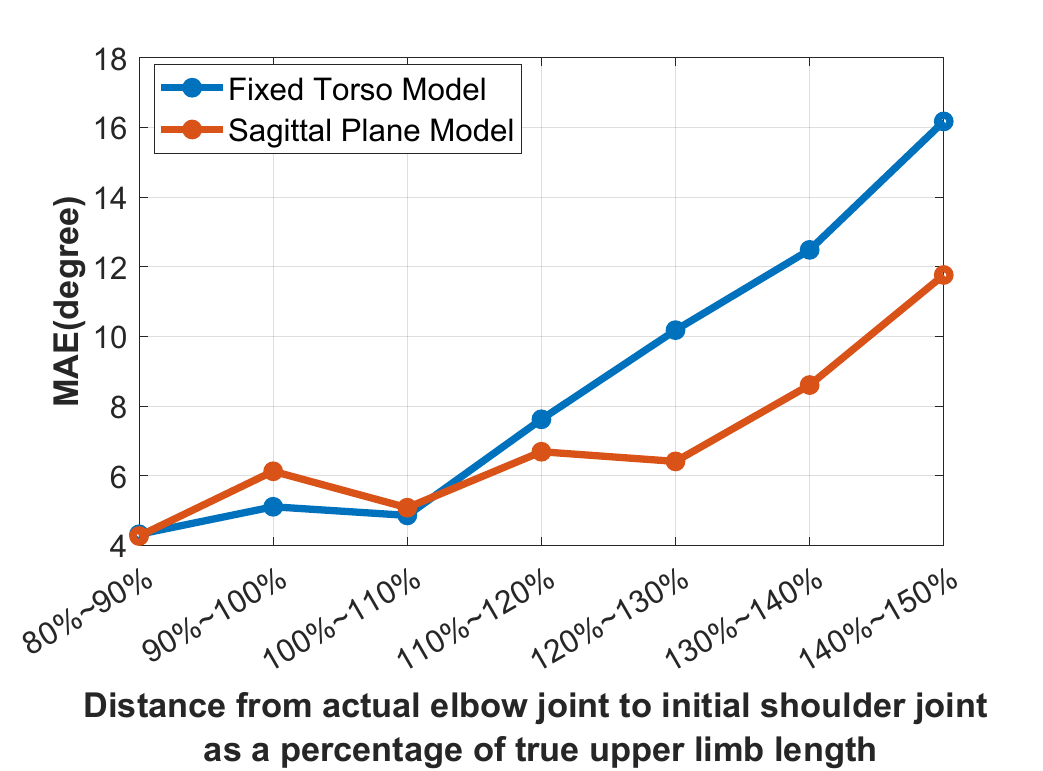}
      \caption{Group classification based on the distance from the real-time elbow joint to the calibrated shoulder joint as the percentage of the actual upper limb length.}
      \label{fig:dis_classify}
\end{figure}

\begin{figure*}[!t]
    \centering
\includegraphics[width=0.9\textwidth]{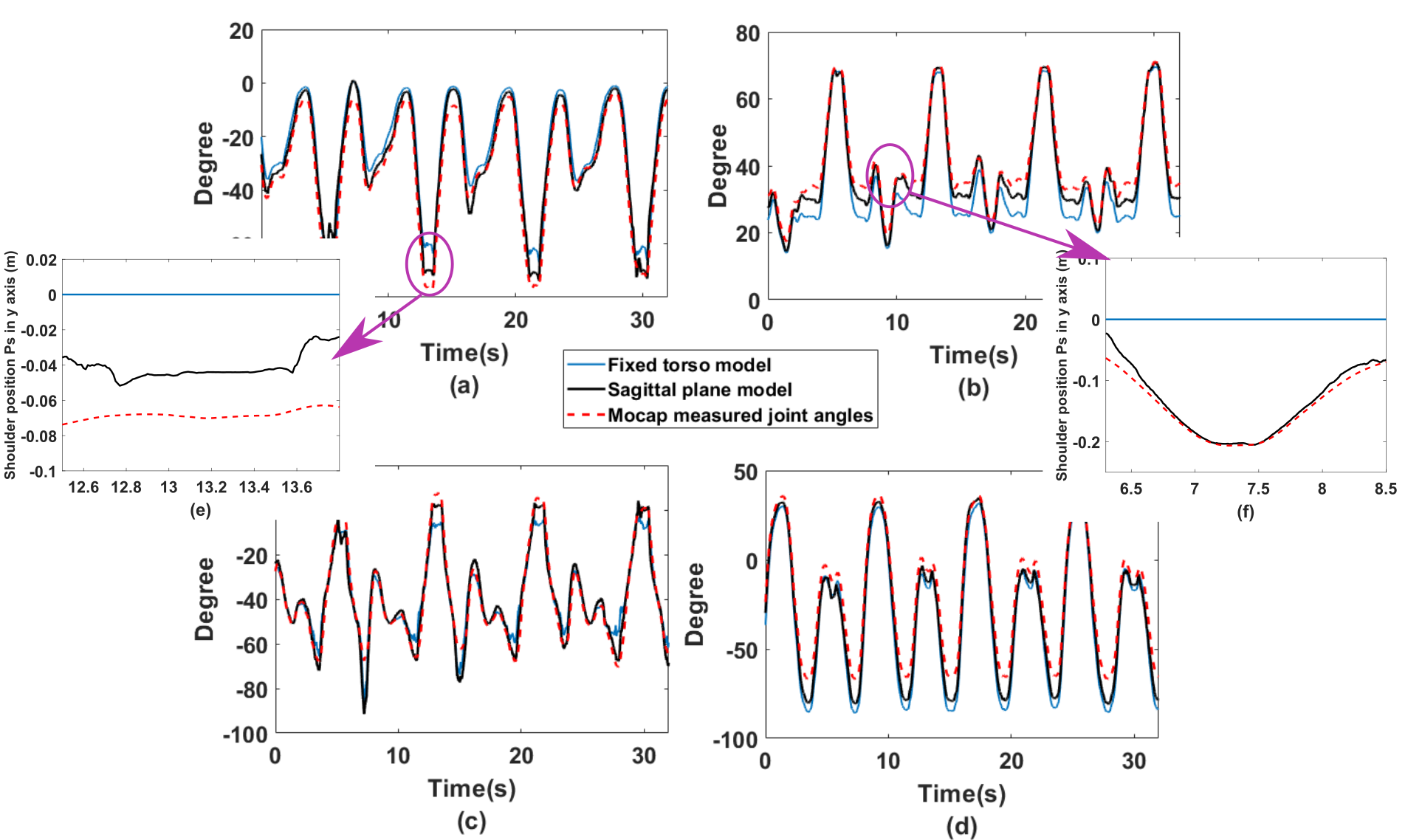}
    \caption{Comparison between Mocap measured angles and estimated joint angles derived from the \textbf{Fixed torso model} and \textbf{Sagittal plane model}. Results were obtained from the label $3$ position of Subject $4$, which demonstrated the change of four joint angles in $30$ seconds. (a) Shoulder Abduction/Adduction.(b) Shoulder Flexion/Extension. (c) Shoulder Internal/External rotation. (d) Elbow Flexion. (e) and (f) The derived shoulder position in the y-axis, which corresponds to the data at the circled position in (a) and (b). }
\label{fig:angle}
\end{figure*}

We first analyzed the impact of distance from the target location on the joint angle estimation performance of two models. The experimental data from all subjects ($5*6*4 = 120$ motions) were classified into $7$ groups. The criteria for group assignment were based on the real-time elbow-to-initial shoulder joint (determined from the initial calibration Sec.\ref{subsec:cal}) distance as a percentage of the actual length of upper arm $l_U$ (fixed value refer to each subject). The percentage from $100\%$ to $150\%$ indicated that the subject's torso moved forward in the sagittal plane and towards the experimental table. The group with $80\%$ to $100\%$ illustrated that the subject's torso moved backward in the sagittal plane and away from the table during experiments. As shown in Fig.\ref{fig:dis_classify}, with the increment of percentage from $100\%$, the joint angle estimation MAE of the fixed torso model dramatically increases from $4.86\degree$ to $16.17\degree$. However, the MAE of the Sagittal plane model rises slowly from $5.08\degree$ to $11.76\degree$. The paired t-test was applied for statistical analysis. There were significant differences in the estimated performance of the two models in terms of the last three groups ($p=0.04, 0.007$, and $0.028$). As illustrated in Exp1, the label $3$ position was designed for evaluating the sagittal plane model. Therefore, we plotted the angle estimation results for the label $3$ of subject $4$ (Fig. \ref{fig:angle}) to compare the changes between the estimated angles of two models and Mocap measured angles (Ground truth). As can be observed in the subfigure (e) and (f), the sagittal plane model can derive a more accurate shoulder position $P_{s}$, which in turn will result in relatively precise estimated joint angles when the torso has significant movements, especially in the case of an abrupt change in the slope of the angle.\par

\begin{figure}[thpb]
\centering
   \includegraphics[scale=0.4]{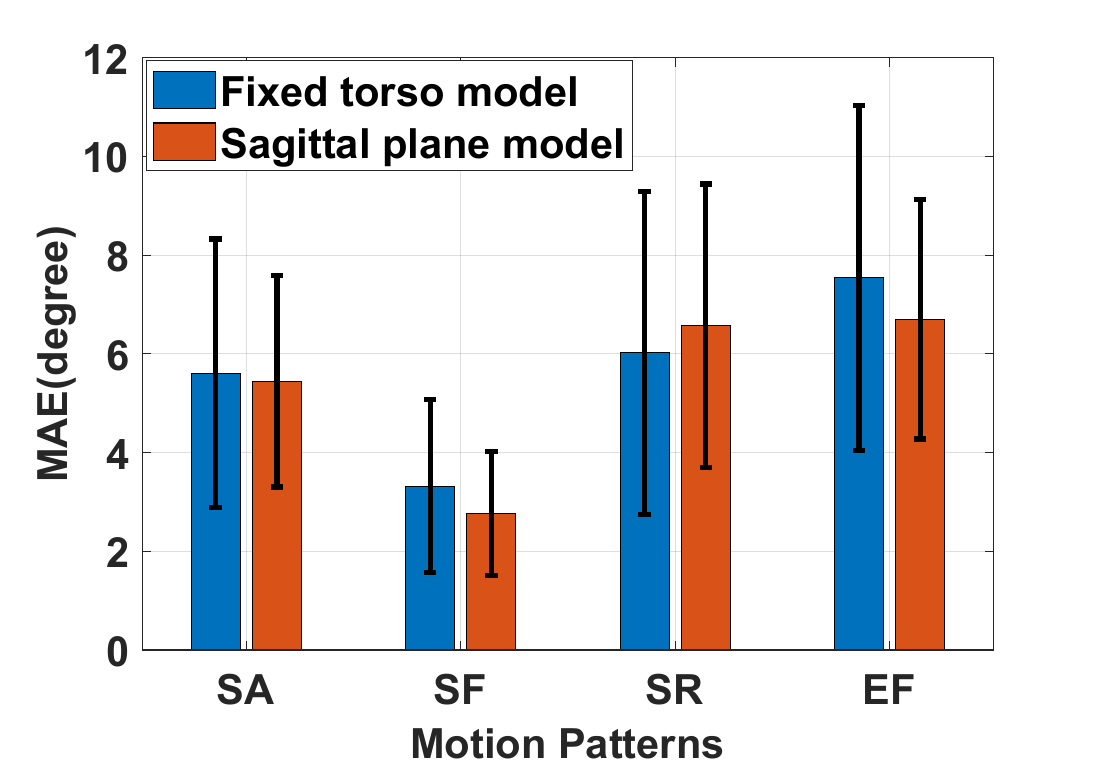}
    \caption{Comparison of two model performances on four types of motion patterns.}
    \label{fig:comp_mae_angle}
\end{figure}

To validate the estimated performance of both models across various types of joint angles, we examined four distinct categories of joint angles referring to MAE and standard derivation (Fig.\ref{fig:comp_mae_angle}). Both models exhibit significantly outstanding estimation performance in shoulder flexion/extension compared to other motion patterns, which are $3.32\degree\pm1.75\degree$ and $2.77\degree\pm1.25\degree$, respectively. However, there are no significant differences between the two proposed models referring to each motion type.
\subsection{Effects of Support Force on Human Arm}
\begin{figure}[thpb]
\centering
\subfigure[]{\label{fig:EMG} \includegraphics[width=0.48\textwidth]{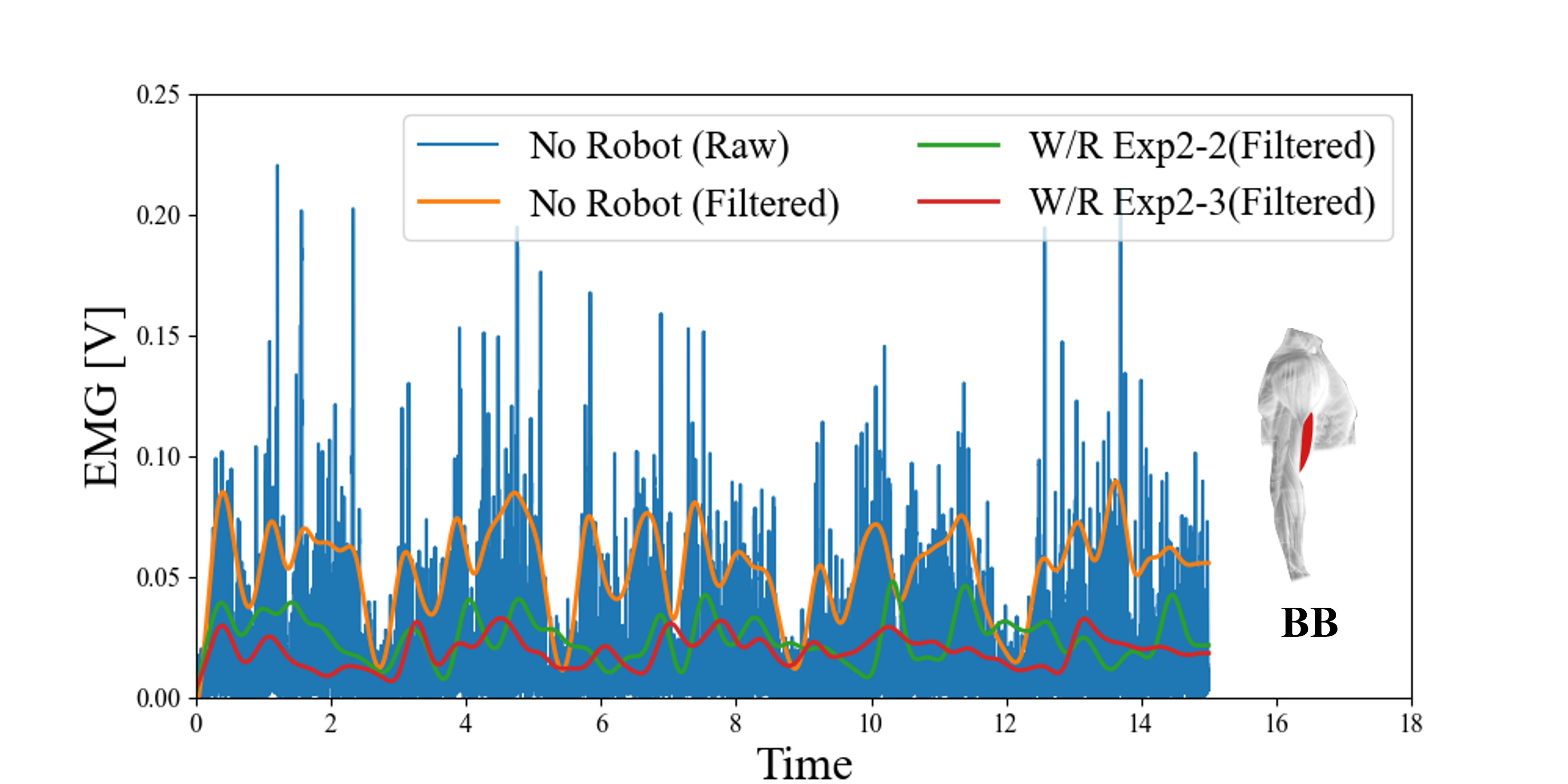}}
\hfill
\subfigure[]{\label{fig:FT} \includegraphics[width=0.48\textwidth]{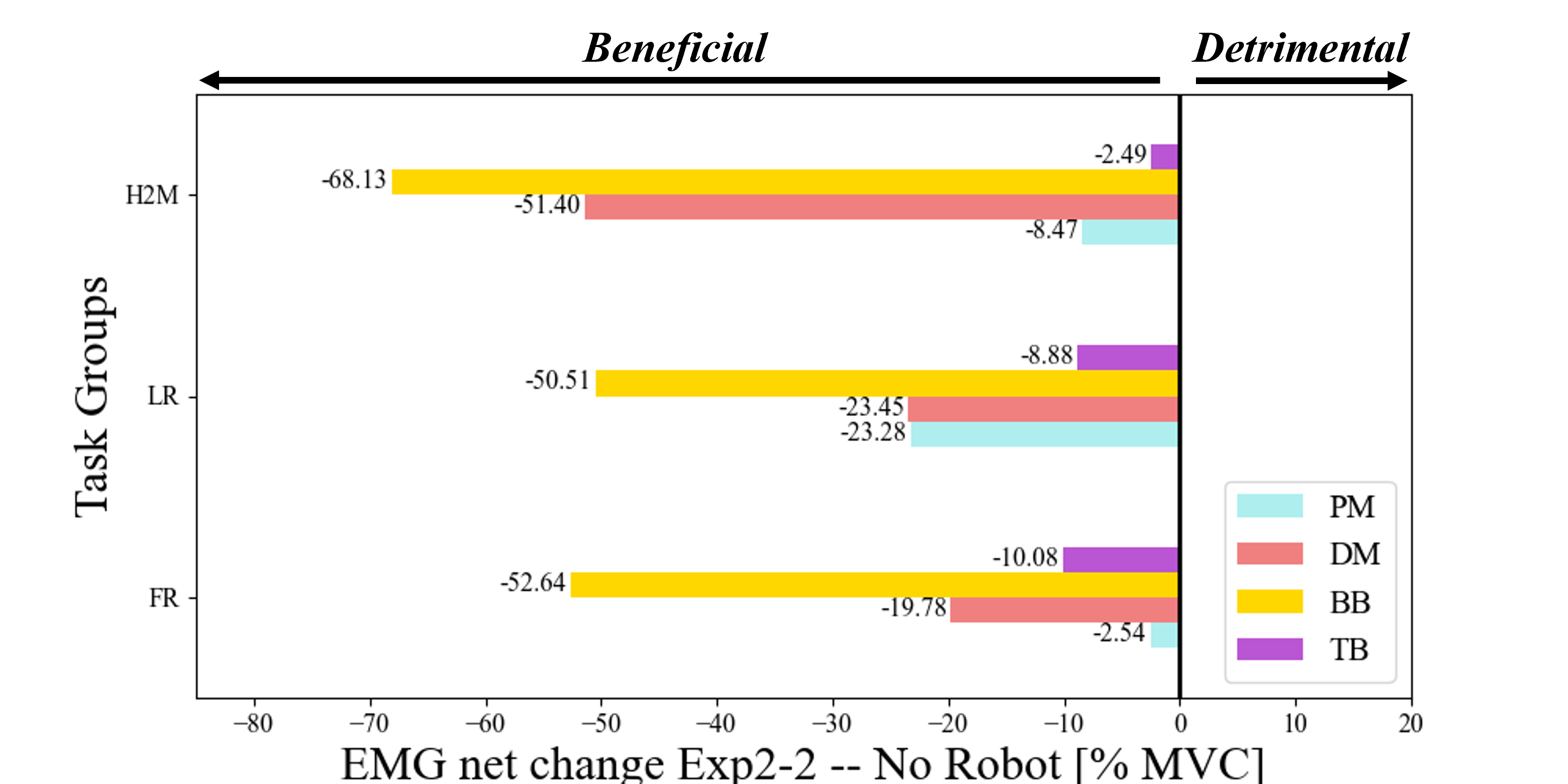}}%
\hfill
\subfigure[]{\label{fig:SP} \includegraphics[width=0.5\textwidth]{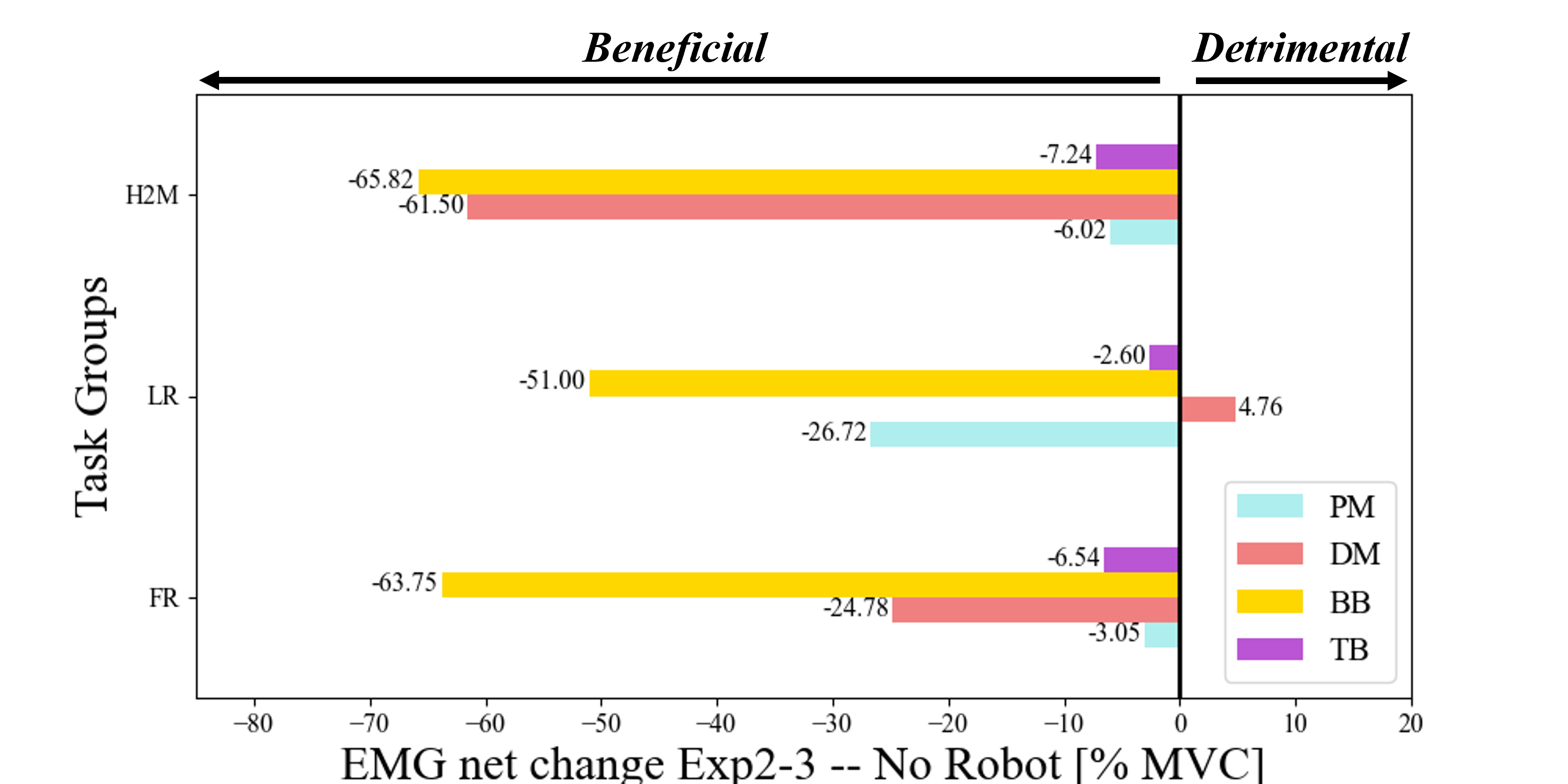}}%
\caption{Effects of the robot support force on human arm. (a). An example of the sEMG profile for the Biceps Brachii (BB) in one subject while performing the 'Forward Reaching' task. (b) and (c) illustrate the net changes in EMG for four muscle activation when transitioning from No Robot mode to With Robot mode, under the fixed torso model (Exp2-2) and the sagittal plane model (Exp2-3), respectively. These changes are differentiated for each task, including H2M, LR, and FR.}
\label{EMG_results}
\end{figure}

In Exp2, we evaluated the effects of the calculated support force by analyzing muscle activities among three experimental sessions. As shown in Fig. \ref{fig:EMG}, a represented subject's raw and filtered BB sEMG signals were recorded among different experimental modes. The filtered envelope of the raw signals, shown as an orange solid curve, corresponds to the 'No Robot' condition, with the raw signals depicted by the blue line. The green and red solid lines respectively represent Exp2-2 and Exp2-3, each conducted under one of the two proposed assistive control strategies. When participants wore the robot and used the proposed assistive control strategy, muscle activity was significantly reduced relative to not wearing the robot. The net change in EMG activity of the $4$ muscles per task can be seen in Fig. \ref{fig:FT} and Fig. \ref{fig:SP}, where each color bar represents the specific muscle. The transition from No Robot to With Robot mode, under the fixed torso model (Exp2-2) and the sagittal plane model (Exp2-3), resulted in an average muscular activation reduction as follows: it was $11.43 \pm 8.72\%$ and $11.93 \pm 10.53\%$ for PM; $31.54 \pm 14.12\%$ and $27.17 \pm 27.10\%$ for the DM; $57.09 \pm 7.85\%$ and $60.18 \pm 6.55\%$ for the BB; $7.15 \pm 3.33\%$ and $5.50 \pm 2.04\%$ for the TB. 
\section{DISCUSSIONS}
\label{sec:discussion}
The ARAE robot is compact, portable, and easy to set up, which is suitable for home-based therapy. Moreover, the proposed adaptive arm support control framework can provide the support force with different human arm poses, offering simple implementation and adaptability to diverse users. 
\subsection{Evaluation of estimated human joint angles}
For comparison of the two proposed angle estimation methods (Exp1), the sagittal plane model is more suitable for estimating joint angles during torso movements. When the shoulder undergoes significant movements in the sagittal plane (Fig. \ref{fig:dis_classify}), the accuracy of both models decreases as the torso moves forward. However, the sagittal plane model significantly improves the angle estimation accuracy compared to the fixed torso model. This enhancement can make our entire framework more universally applicable, particularly during torso movements. Most importantly, numerous individuals who have suffered from strokes display an excessive use of compensatory trunk motions while reaching and placing objects, which affect the recovery in stroke patients \cite{cirstea2000compensatory}. Therefore, our proposed method can provide accurate joint angle estimation, subsequently enabling the generation of sufficiently precise gravity compensation. Thus, avoiding the trunk compensatory or torso movements improves upper extremity recovery in stroke patients. \par 
Another hypothesis posits a correlation between the estimation performance of the proposed models and the specific type of motion patterns. As the analysis of the results points out, both models show a significant MAE decrease for the SF angle. This is due to the SF angle only correlates with the elbow position in z-direction ${ }^{S}\textbf{p}_{E}(z)$ and $l_{U_{cal}}$. This demonstrates the capability to predict the elbow joint position with relative accuracy solely based on robot joint position information. Moreover, the estimations of the two proposed models are not significantly different for different joint angles. This may be attributed to the fact that the sagittal plane model is primarily sensitive to large torso movements. However, significant torso movement occurs only when the subject reaches forward to label $3$, and this motion constitutes a relatively small percentage of the entire dataset. Consequently, this results in no significant difference in the performance of the two models in estimating joint angles within the context of the overall analysis.
Moreover, the benefit of the proposed models provides the capability to monitor the patient's arm posture during ADL, which is a main feature for clinical assessments of the upper limb \cite{lee2017automated}. There are some attempts to use external sensors to measure the upper limb posture when using the 3D end-effector type of rehabilitation robot, such as magnetic sensors and IMUs \cite{crocher2018upper}, or external RGB-D cameras \cite{hwang2020novel}. However, our proposed methods do not need to rely on external sensors, which reduce the system complexity and enhance the robotic system's usability. 
\subsection{Benefits of Using the ARAE Robot}
The muscular activities of the upper limb were not evaluated in the assistive control mode of EMU and Burt. As for the ARAE robot, one of the observed benefits is the reduction of muscular activity of healthy subjects during simulated ADLs (Exp2). As shown in Fig.\ref{EMG_results}, the BB activity significantly reduces compared with other muscles. 
In the Forward Reaching (FR) task, a notable decrease in muscle activity was observed: for the Biceps Brachii (BB), the reduction ranged from $-52.64\%$ to $-63.75\%$, and for the Deltoid Muscle (DM), it went from $-19.78\%$ to $-24.78\%$ when the control mode utilizing the sagittal plane model (Exp2-3) was implemented. The main reason for this effect is that subjects do trunk compensatory maneuvers during the FR task because position 3 is located in the farthest position. Therefore, the sagittal plane model is better able to derive the changing shoulder joint position, thus obtaining more accurate joint angle prediction. In turn, it provides more precise arm support during FR tasks. However, from inspection of Fig. \ref{fig:SP}, the DM activity has an increase for the LR task when conducting the sagittal plane mode (Exp2-3). This result hypothesizes that the sagittal plane model possesses relatively weak generalization ability when the torso is moving in the later plane. Therefore, the calculated force based on the proposed control framework might produce extra assistive force in some posture to restrict the arm movement during LR task.

\subsection{Limitations}
The adaptive arm gravity compensation control framework exhibits several remaining limitations. Firstly, the model's input variables, including the positions of the shoulders or the center of the pelvis to the robot base, as well as the anthropometric parameters of the patients, necessitate pre-definition before the practical implementation of ARAE. Consequently, the introduction of a personalized model becomes imperative to address this concern. Secondly, an expanded participant pool involving stroke patients is essential for evaluating the robustness and effective of the system. 

\section{CONCLUSIONS}
\label{sec:con}
The 3D end-effector type of rehabilitation robot ARAE presented in this work is a versatile and compact robotic system, accommodating a wide range of patients with Muscle Manual Testing (MMT) scores from 1 to 4. When combined with the proposed adaptive gravity compensation control framework, it shows promise in enhancing the applicability of the device for home-based physical therapy. This includes facilitating ADL tasks and interaction with practical environments. The proposed adaptive gravity compensation control framework enables to provide adaptive arm support in three-dimensional space, which does not rely on wearable sensors to measure human arm postures. Experiments has unveiled significant enhancements in angle estimation accuracy within the sagittal plane model, particularly evident when accommodating substantial movements of the torso. The contrast against the fixed torso mode underscores the pivotal role of adaptability in optimizing performance. Moreover, the experiments of no/with robot were performed to evaluate the effects on the human body. \par
Future work will continue to develop the personalized model, thus ensuring that the framework can be adaptive to more tasks and different subjects. Furthermore, patient trials need to be conducted to further assess the system's usability and the performance of the proposed control framework.












\bibliographystyle{IEEEtran}
\bibliography{references}

\end{document}


\title{Supplementary Material for Design and Evaluation of a Compact 3D End-effector Assistive Robot for Adaptive Arm Support}

\author{Sibo Yang, Lincong Luo, Wei Chuan Law, Youlong Wang, Lei Li and Wei Tech Ang}



\maketitle

\renewcommand{\thesection}{S.\Roman{section}} 
\renewcommand{\thesubsection}{\thesection.\Alph{subsection}}
\section{The process of forward kinematics model}
\begin{figure}[!htbp]
  \centering
  \includegraphics[width=1\linewidth]{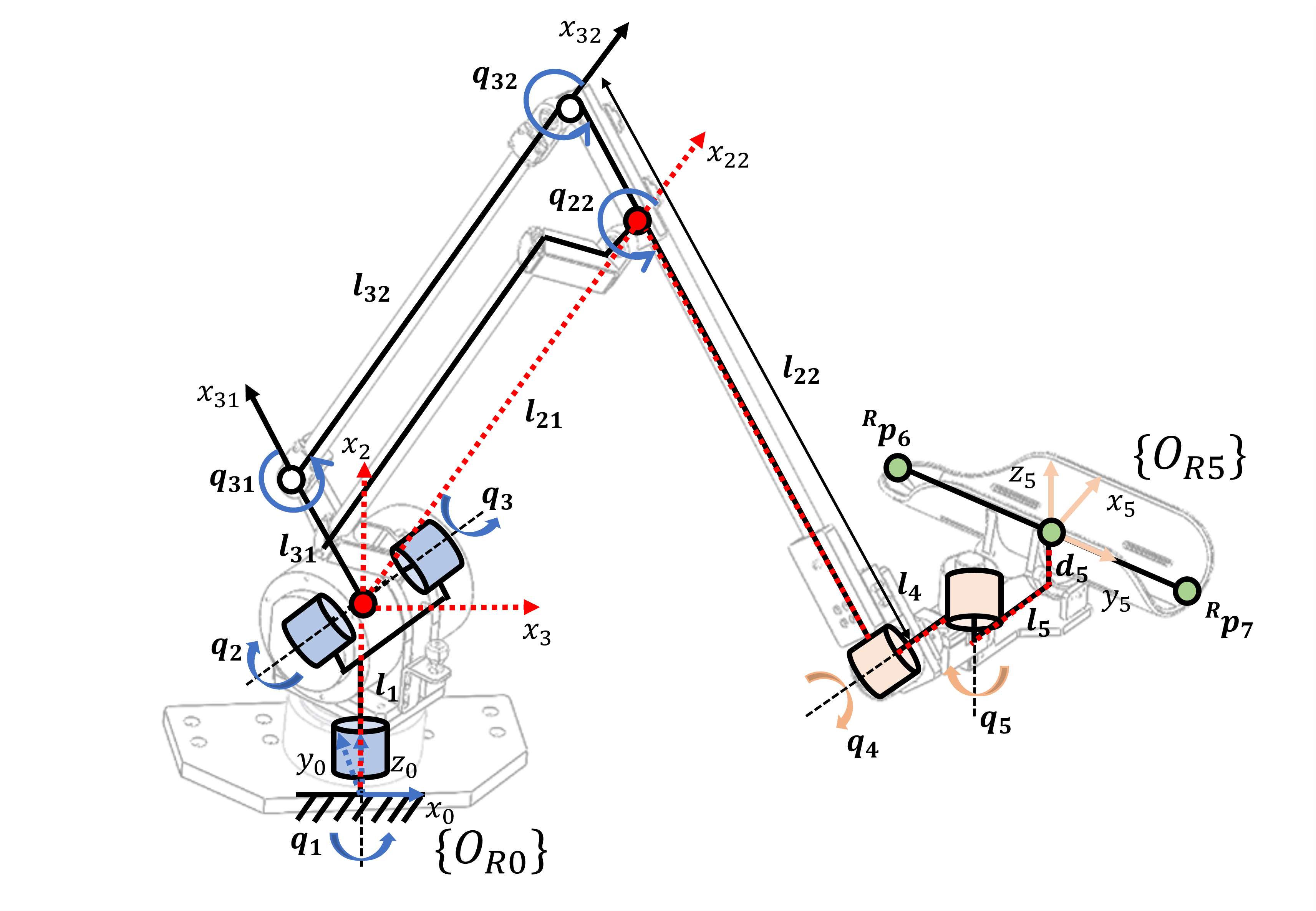}
  \caption{Schematic of the kinematics model of ARAE. The green circles demonstrate the interaction points between humans and robots.}
  \label{fig:FK}
\end{figure}

As shown in Fig. \ref{fig:FK}, the joint angles and structural parameters are assigned to each joint and link. Since the main drive mechanism of ARAE is a parallelogram structure, the kinematic model can be simplified as follows serial-link mechanism, clarified as the red dashed line in Fig.\ref{fig:FK}. Therefore, the Denavit-Hartenberg (D-H) algorithm is applied to derive the kinematics model. \par
\subsection{Denavit-Hartenberg (DH) parameters}
The Denavit-Hartenberg (D-H) algorithm is applied to derive the kinematics model of ARAE. The D-H parameters are defined in the following Table \ref{tab:D-H para}. In this table, $q_{22}$ can be articulated as a function of $q_{3}$ (Eq. \ref{equ:q22}), leveraging the properties of the parallelogram mechanism. Moreover, the length of links is shown in Table.\ref{tab:link_para}. 
\begin{equation}
q_{22} = q_{3} - (\frac{\pi}{2} - q_{2}) + \pi
\label{equ:q22}
\end{equation}
\begin{table}[h]
    \centering
    \caption{D-H parameters for ARAE}
    \begin{tabular}{c|cccc}
     joint i & $q_i$ & $\alpha_{i}$ & $a_{i}$ & $d_i$\\ \hline
     1  & $q_1$ & $\frac{\pi}{2}$ & 0 & $l_1$\\ \hline
     2  & $\frac{\pi}{2} - q_{2}$ & 0 & $l_{21}$ & 0\\ \hline
     3  & $q_{22}$ & 0 & $l_{22} - l_{31}$ & 0\\ \hline
     4  & $q_4$ & $-\frac{\pi}{2}$ & 0 & $-l_4$\\ \hline
     5  & $q_5$ & 0 & $l_5$ & $d_5$\\ \hline
    \end{tabular}
    \label{tab:D-H para}
\end{table}
\begin{table}[h]
    \centering
    \caption{The link parameters of ARAE}
    \begin{tabular}{|l|l|l|l|}
    \hline
    $l_1$     & $l_{21}$    & $l_{2}$     & $l_{31}$   \\ \hline
    0.068m & 0.43m  & 0.446m  & 0.1m  \\ \hline
    $l_{32}$    & $l_4$     & $l_5$      & $d_5$    \\ \hline
    0.43m  & 0.111m & 0.0895m & 0.01m \\ \hline
    \end{tabular}
    \label{tab:link_para}
\end{table}
\subsection{Forward Kinematics model}
The defined D-H parameters are input to the homogeneous transformation matrix $\left[{ }^{i-1} \textbf{T}_i\right]$ (denoted as Eq. \ref{equ:DH matrix}), which denotes the D-H matrix relating to frame $i-1$ and frame $i$. 
\begin{equation}
\left[{ }^{i-1} \textbf{T}_i\right]=\left[\begin{array}{cccc}
c q_i & -c \alpha_i s q_i & s \alpha_i s q_i & a_i c q_i \\
s q_i & c \alpha_i c q_i & -s \alpha_i c q_i & a_i s q_i \\
0 & s \alpha_i & c \alpha_i & d_i \\
0 & 0 & 0 & 1
\end{array}\right]
\label{equ:DH matrix}
\end{equation}
The $c$ is $\cos$ and $s$ is $\sin$. Then, the equation of the resultant matrix ${}^{0}\textbf{T}_{5}$ specifies the fifth frame to the base coordinate of the robot following the chain product which is denoted as: 
\begin{equation}
{}^{0}\textbf{T}_{5}  = { }^{0}\textbf{T}_{1}\cdot{ }^{1} \textbf{T}_{2}\cdot{ }^{2} \textbf{T}_{3}\cdot{ }^{3} \textbf{T}_{4}\cdot{ }^{4} \textbf{T}_{5}
\label{equ:rel_mat}
\end{equation}
The vector of end-effector position is ${ }^{R}\textbf{p}_{5} \in \mathbb{R}^{4 \times 1}$ under robot based frame, which follows the Equation.\ref{equ:FK_robot}. The two endpoints of the cuff are denoted as ${ }^{R}\textbf{p}_{6}$ and ${ }^{R}\textbf{p}_{7}$ in the robot base coordinate system, which can be defined as:
\begin{equation}
    { }^{R}\textbf{p}_{5} = { }^{0}\textbf{p}_{5} = {}^{0}\textbf{T}_{5}\cdot \textbf{p}_{5}
    \label{equ:FK_robot}
\end{equation}
\begin{equation}
    { }^{R}\textbf{p}_{6} = { }^{0}\textbf{p}_{6} = {}^{0}\textbf{T}_{5}\cdot {}^{5}\textbf{T}_{6} \cdot\textbf{p}_{6}
    \label{equ:P6}
\end{equation}
\begin{equation}
    { }^{R}\textbf{p}_{7} = { }^{0}\textbf{p}_{7} = {}^{0}\textbf{T}_{5}\cdot {}^{5}\textbf{T}_{6} \cdot {}^{6}\textbf{T}_{7}\cdot\textbf{p}_{7}
    \label{equ:P7}
\end{equation}
where $\textbf{p}_{5}, \textbf{p}_{6} $ and $\textbf{p}_{7}$ represent the 5th, 6th, and 7th joint positions, respectively, in the specific local coordinates. The transformation matrices ${}^{5}\textbf{T}_{6}$ and ${}^{6}\textbf{T}_{7}$ represent the transformations from the 5th joint to the 6th joint and from the 6th joint to 7th joint, respectively. \par

\subsection{Robot jacobian matrix}
The robot Jacobin matrix is $\textbf{J}_{R} \in \mathbb{R}^{3 \times 3}$. Individual elements $\textbf{J}_{R}$ consist of as following Equation:
\begin{equation}
\centering
\textbf{J}_{r} = 
\left[\begin{array}{ccc}
A_{11} & A_{12} & A_{13} \\
A_{21} & A_{22} & A_{23} \\
0 & A_{32} & A_{33}
\end{array}\right]
\label{equ:jacobian}
\end{equation}
where
\begin{equation}
    \begin{aligned}
    &A_{11} = -l_{21} \sigma_1 \sin(q_1) - \cos(\beta) \sigma_1 \sin(q_1) (l_{22} - l_{31}) \\
    &- \sin(\beta) \sin(q_1) \sigma_2 (l_{22} - l_{31}) \\
    \end{aligned}
\end{equation}
\begin{equation}
\begin{aligned}
    &A_{12} = -l_{21} \cos(q_1) \sigma_2 \\
\end{aligned}
\end{equation}
\begin{equation}
\begin{aligned}
    &A_{13} = \cos(\beta) \cos(q_1) \sigma_2 (l_{22} - l_{31}) \\
    &- \sin(\beta) \cos(q_1) \sigma_1 (l_{22} - l_{31}) \\
\end{aligned}
\end{equation}

\begin{equation}
\begin{aligned}
    &A_{21} = l_{21} \cos(q_1) \sigma_1 + \cos(\beta) \cos(q_1) \sigma_1 (l_{22} - l_{31}) \\
    &+ \sin(\beta) \cos(q_1) \sigma_2 (l_{22} - l_{31}) \\
\end{aligned}
\end{equation}
   
\begin{equation}
\begin{aligned}
&A_{22} = -l_{21} \sin(q_1) \sigma_2
\end{aligned}
\end{equation}

\begin{equation}
\begin{aligned}
    &A_{23} = \cos(\beta) \sin(q_1) \sigma_2 (l_{22} - l_{31}) \\
    &- \sin(\beta) \sigma_1 \sin(q_1) (l_{22} - l_{31}) \\
\end{aligned}
\end{equation}    

\begin{equation}
\begin{aligned}
&A_{32} = -l_{21} \sigma_1 \\
\end{aligned}
\end{equation}

\begin{equation}
\begin{aligned}
&A_{33} = \cos(\beta) \sigma_1 (l_{22} - l_{31}) + \sin(\beta) \sigma_2 (l_{22} - l_{31}) \\
\end{aligned}
\end{equation}
where 
\begin{equation}
    \left\{
    \begin{aligned}
    &\sigma_1 = \cos\left(q_{2} -\frac{\pi}{2}\right), \\
    &\sigma_2 = \sin\left(q_{2} -\frac{\pi}{2}\right), \\
    &\beta = q_{2} + q_{3} + \frac{\pi}{2}.
    \end{aligned}
\right.
\end{equation}

\section{Expression of inverse kinematics model}
The inverse kinematics (IK) model is constrained to only compute the active joint angles of the robot using the \({ }^{R}\mathbf{p}_{3}\) input. The \({ }^{R}\mathbf{p}_{3}\) is denoted as $\left[x, y, z\right]^{T}$.\par
The following equations are used to solve three active joints with the input of $\left[x, y, z\right]^{T}$:
\begin{equation}
q_1 = \arctan\left(\frac{y}{x}\right)
\end{equation}
The cosine rule is applied to calculate \(q_2\):
\begin{equation}
\left\{
\begin{aligned}
&\delta = (z - l_1)^2 + x^2 + y^2, \\
&\cos{a} = \frac{l_{21}^2 + \delta - (l_{22} - l_{31})^2}{2 \cdot l_{21} \cdot \sqrt{\delta}},\\
&\sin{a} = \sqrt{1 - (\cos{a})^2},\\
&q_{2} = \frac{\pi}{2} - \left(\arctan\left(\frac{z - l_1}{\sqrt{x^2 + y^2}}\right) + \arctan\left(\frac{\sin{a}}{\cos{a}}\right)\right),\\
\end{aligned}
\right.
\end{equation}
Then, the joint angle $q_3$ can be obtained as follow:
\begin{equation}
\left\{
    \begin{aligned}
        &\cos(q_{22}) = \frac{\delta - l_{21}^2 - (l_{22} - l_{31})^2}{2 \cdot l_{21} \cdot (l_{22} - l_{31})},\\
        &\sin(q_{22}) = \sqrt{1 - (\cos(q_{22}))^2},\\
        &q_{22} = - \arctan\left(\frac{\sin(q_{22})}{\cos(q_{22})}\right),\\
        &q_3 = \left(\arctan\left(\frac{z - l_1}{\sqrt{x^2 + y^2}}\right) + \arctan\left(\frac{\sin{a}}{\cos{a}}\right)\right) \\
        &\qquad - \pi + q_{22},
    \end{aligned}
\right.
\end{equation}
\section{Human arm dynamics model}
The human arm dynamics model can be written as:
\begin{equation}
\label{equ:human DY}
\mathbf{M}_h\left(h_j\right) \ddot{h}_j+\mathbf{C}_h\left(h_j, \dot{h}_j\right) \dot{h}_j+\mathbf{G}_h\left(h_j\right)=\mathbf{\tau}
\end{equation}
where $h_j$, $\dot{h}_j$ and $\ddot{h}_j \in \mathbb{R}^n$ are the human arm joint angles and derivatives. $\mathbf{\tau}$ is the joint torque generated by the human arm. $n = 4$ denotes the four DOFs of the human arm model including three DOFs in the shoulder joint and one DOF in the elbow joint. \par
The robot applies a support force $\mathbf{F}_h$ to the human arm, forming an external torque $\mathbf{R}_r$ that primarily compensates for the arm's gravity term $\mathbf{G}_h$, as illustrated in Equation \ref{equ:updated human DY}.
\begin{equation}
\label{equ:updated human DY}
\mathbf{M}_h\left(h_j\right) \ddot{h}_j+\mathbf{C}_h\left(h_j, \dot{h}_j\right) \dot{h}_j+\mathbf{G}_h\left(h_j\right)=\mathbf{\tau} + \mathbf{R}_r\left(h_{j},\mathbf{F}_{h}\right)
\end{equation}
Therefore, the external torque $R_r$ equals to the human arm gravity term $\mathbf{G}_h$ can be denoted as:
\begin{equation}
    \mathbf{R}_r\left(h_{j},\mathbf{F}_{h}\right) = \mathbf{G}_h
\end{equation}
Therefore, the required force of the end-effector to support the human arm's weight can be expressed by the human arm gravity term $\mathbf{G}_h$:
\begin{equation}
    \mathbf{F}_h = \mathbf{J}_h^{T \#}\left(\mathbf{h}_j\right) \mathbf{G}_h\left(\mathbf{h}_j\right)
\label{equ:hd2}
\end{equation}
The detailed expression of the jacobian matrix of human arm $\mathbf{J}_h\left(h_j\right)$ is as follows:
\begin{equation}
\centering
\mathbf{J}_h\left(h_j\right) = 
\left[\begin{array}{cccc}
B_{11} & B_{12} & B_{13} & B_{14}\\
B_{21} & B_{22} & B_{23} & B_{24}\\
0 & B_{32} & B_{33} & B_{34}
\end{array}\right]
\label{equ:jacobian_human}
\end{equation}
where
\begin{equation}
    \begin{aligned}
    &B_{11} = l_{U}\,\cos \left(h_1 \right)\,\cos \left(h_2 \right)-\frac{l_{F}}{2} \,\cos \left(h_4 \right)\,\lambda_2 \\
    &-\frac{l_{F}}{2} \,\cos \left(h_1 \right)\,\cos \left(h_2 \right)\,\sin \left(h_4 \right)
    \end{aligned}
\end{equation}
\begin{equation}
    \begin{aligned}
    &B_{12} = \frac{l_{F}}{2} \,\sin \left(h_1 \right)\,\sin \left(h_2 \right)\,\sin \left(h_4 \right)-l_{U}\,\sin \left(h_1 \right)\,\sin \left(h_2 \right) \\
    &-\frac{l_{F}}{2} \,\cos \left(h_2 \right)\,\cos \left(h_4 \right)\,\sin \left(h_1 \right)\,\sin \left(h_3 \right)
    \end{aligned}
\end{equation}
\begin{equation}
    \begin{aligned}
    &B_{13} = -\frac{l_{F}}{2} \,\cos \left(h_4 \right)\,{\left(\cos \left(h_1 \right)\,\sin \left(h_3 \right)+\cos \left(h_3 \right)\,\sin \left(h_1 \right)\,\sin \left(h_2 \right)\right)}
    \end{aligned}
\end{equation}
\begin{equation}
    \begin{aligned}
    &B_{14} = -\frac{l_{F}}{2} \,\sin \left(h_4 \right)\,\sigma_1 -\frac{l_{F}}{2} \,\cos \left(h_2 \right)\,\cos \left(h_4 \right)\,\sin \left(h_1 \right)
    \end{aligned}
\end{equation}
\begin{equation}
    \begin{aligned}
    &B_{21} = l_{U}\,\cos \left(h_2 \right)\,\sin \left(h_1 \right)+\frac{l_{F}}{2} \,\cos \left(h_4 \right)\,\lambda_1 \\
    &-\frac{l_{F}}{2} \,\cos \left(h_2 \right)\,\sin \left(h_1 \right)\,\sin \left(h_4 \right)
    \end{aligned}
\end{equation}
\begin{equation}
    \begin{aligned}
    &B_{22} = l_{U}\,\cos \left(h_1 \right)\,\sin \left(h_2 \right)-\frac{l_{F}}{2} \,\cos \left(h_1 \right)\,\sin \left(h_2 \right)\,\sin \left(h_4 \right) \\
    &+\frac{l_{F}}{2} \,\cos \left(h_1 \right)\,\cos \left(h_2 \right)\,\cos \left(h_4 \right)\,\sin \left(h_3 \right)
    \end{aligned}
\end{equation}
\begin{equation}
    \begin{aligned}
    &B_{23} =  -\frac{l_{F}}{2} \,\cos \left(h_4 \right)\,{\left(\sin \left(h_1 \right)\,\sin \left(h_3 \right)+\cos \left(h_1 \right)\,\cos \left(h_3 \right)\,\sin \left(h_2 \right)\right)}
    \end{aligned}
\end{equation}
\begin{equation}
    \begin{aligned}
    &B_{24} =  \frac{l_{F}}{2} \,\cos \left(h_1 \right)\,\cos \left(h_2 \right)\,\cos \left(h_4 \right)-\frac{l_{F}}{2} \,\sin \left(h_4 \right)\,\lambda_2
    \end{aligned}
\end{equation}

\begin{equation}
    \begin{aligned}
    &B_{32} =  \frac{l_{F}}{2} \,\cos \left(h_2 \right)\,\sin \left(h_4 \right)-l_{U}\,\cos \left(h_2 \right) \\
    &+\frac{l_{F}}{2} \,\cos \left(h_4 \right)\,\sin \left(h_2 \right)\,\sin \left(h_3 \right)
    \end{aligned}
\end{equation}

\begin{equation}
    \begin{aligned}
    &B_{33} =  -\frac{l_{F}}{2} \,\cos \left(h_2 \right)\,\cos \left(h_3 \right)\,\cos \left(h_4 \right)
    \end{aligned}
\end{equation}

\begin{equation}
    \begin{aligned}
    &B_{34} =  \frac{l_{F}}{2} \,\cos \left(h_4 \right)\,\sin \left(h_2 \right)+\frac{l_{F}}{2} \,\cos \left(h_2 \right)\,\sin \left(h_3 \right)\,\sin \left(h_4 \right)
    \end{aligned}
\end{equation}
where

$\begin{array}{l}
\mathrm{}\\
\;\;\lambda_1 =\cos \left(h_1 \right)\,\cos \left(h_3 \right)-\sin \left(h_1 \right)\,\sin \left(h_2 \right)\,\sin \left(h_3 \right)\\
\mathrm{}\\
\;\;\lambda_2 =\cos \left(h_3 \right)\,\sin \left(h_1 \right)+\cos \left(h_1 \right)\,\sin \left(h_2 \right)\,\sin \left(h_3 \right)
\end{array}$

The $l_U$ and $l_F$ represent the length of upper limb and forearm. In addition, the human arm gravity vector $\mathbf{G}_h\left(\mathbf{h}_j\right)$ refers to:
\begin{equation}
\centering
\mathbf{G}_h\left(\mathbf{h}_j\right) = 
\left[\begin{array}{c}
0\\
G_{21}\\
-l_{F}\cdot COM_{F}\cdot m_{F} \cos \left(h_2\right) \cos \left(h_3\right) \cos \left(h_4\right)g\\
G_{41}\\
\end{array}\right]
\label{equ:gravity_human}
\end{equation}

\begin{equation}
    \begin{aligned}
    &G_{21} =  g(m_{L}(l_{F} \cdot COM_{F} \cdot \cos \left(h_2\right) \sin \left(h_4\right) \\
    &-l_{U} \cos \left(h_2\right) l_{F} \cdot COM_{F} \cdot \cos \left(h_4\right) \sin \left(h_2\right) \sin \left(h_3\right)) \\
    &-l_{U} \cdot COM_{U} \cdot m_{U} \cos \left(h_2\right))
    \end{aligned}
\end{equation}

\begin{equation}
    \begin{aligned}
    &G_{41} =  g m_{L}(l_{F} \cdot COM_{F} \cdot \cos \left(h_4\right) \sin \left(h_2\right) \\
    &+l_{F} \cdot COM_{F} \cos \left(h_2\right) \sin \left(h_3\right) \sin \left(h_4\right))
    \end{aligned}
\end{equation}
The $COM_{F}$ and $COM_{U}$ refer to the ratio of center of mass point of forearm and upper limb. Moreover, the $m_U$ and $m_L$ are the mass of upper limb and forearm.

\vfill